\documentclass[fleqn,10pt]{wlscirep}

\graphicspath{{./figures/}{./photos/}{./bio/}}
\usepackage{hyperref}
\usepackage{amsmath,amssymb,amsfonts}
\usepackage{textcomp}
\usepackage{subcaption}
\usepackage{gensymb}
\usepackage{float}
\usepackage{algorithm,algcompatible,algpseudocode}
\usepackage{array}
\usepackage{eso-pic,rotating,graphicx}
\usepackage{multirow}
\usepackage{multicol}
\usepackage{tabularx, makecell}
\usepackage[utf8]{inputenc}
\usepackage{svg}
\usepackage{changebar}  
\usepackage{ctable}
\usepackage{booktabs}
\usepackage{lineno}
\usepackage{threeparttable}
\usepackage[justification=centering]{caption}
\usepackage{soul}
\sethlcolor{yellow}
\usepackage{longtable}
\definecolor{revgreen}{rgb}{0.72,0.96,0.72}

\title{A Network Science Approach to Granular Time Series Segmentation}

\author[1]{Ivana Kesić}
\author[1]{Carolina Fortuna}
\author[1,2]{Mihael Mohor\v{c}i\v{c}}
\author[1*]{Bla\v{z} Bertalani\v{c}}
\affil[1]{Jo\v{z}ef Stefan Institute, SI-1000 Ljubljana}
\affil[2]{Jo\v{z}ef Stefan International Postgraduate School, SI-1000 Ljubljana}
\affil[*]{blaz.bertalanic@ijs.si}

\begin{abstract}

Time series segmentation assigns a label to each part of a sequence. We formulate dense univariate segmentation as node classification on a graph whose nodes are the original time points. A local window provides node features without setting output granularity. We evaluate the approach on a TSSB-derived inductive benchmark built from disjoint UCR training and test instances. Under one fixed Graph Attention Network (GAT), visibility-based transformations achieve the highest mean ranks among eleven graph constructions. WDPVG, directed NVG, and weighted NVG form a statistically indistinguishable top group after Holm correction. On the 59-dataset Time Series Segmentation Benchmark, WDPVG+GAT reaches a weighted F1 of $0.916$, below seq2point at $0.951$ and statistically indistinguishable from same-feature MLP, random-forest, and 1-NN controls, because at this downsampled resolution each segment is short and the fixed $81$-sample window already spans most of it. At native resolution, where each segment is longer than that window, WDPVG+GAT is less sensitive to feature-window width and remains above the same-feature MLP at every tested window. The graph's advantage over these point-wise classifiers comes from context beyond the local window, which the visibility graph's long-range edges can supply, rather than from better features within it. In a separate resolution sweep, it is statistically tied with a window-searched seq2point while using about $70\times$ fewer parameters and $46\times$ less measured peak memory, although seq2point moves ahead after downsampling. This identifies a practical operating regime for finely sampled series when model size and repeated window tuning matter.

\end{abstract}

\begin{document}

\flushbottom
\maketitle
\thispagestyle{empty}


\section{Introduction}

Time series segmentation (TSS) partitions a sequence into regions that correspond to distinct states or activities. Dense segmentation assigns a semantic label to every time point and supports later tasks such as classification, anomaly detection, and process analysis~\cite{carvalho2019systematic,qaddoura2021multi,wang2024unsupervised}. The task is difficult when segment classes differ only in local waveform shape or when relevant context extends beyond a short neighbourhood.

Convolutional, recurrent, and transformer-based models have been used for dense time-series labelling~\cite{perslev2019u,phan_8631195,yao2018efficient}. These models operate on the temporal sequence, often through a finite input window. A graph representation provides a different inductive bias. Each time point becomes a node, and edges encode shape-based relations between points that need not be adjacent in time~\cite{zheng2021visibility,lacasa2008time,luque2009horizontal,supriya2016weighted}. A graph neural network can then assign one label to each time-point node. In our implementation, a local window supplies label-free node features, but it does not determine the spacing or number of output predictions.

We study this formulation on a TSSB-derived inductive benchmark that follows the original TSSB series-generation procedure while separating the UCR instances used for development from those used for final testing. We evaluate eleven time-series-to-graph transformations with one fixed GAT architecture. Visibility-based graphs achieve the highest mean ranks under the tested settings. WDPVG, directed NVG, and weighted NVG form a statistically indistinguishable top group after correction. We use WDPVG for the detailed experiments because its dual perspective represents peaks and troughs symmetrically, not because it is uniquely more accurate or more efficient.
The main contributions are:

\noindent \textbf{1. An inductive graph formulation for dense univariate segmentation.}
    We cast supervised univariate time-series segmentation as node classification
    with one node and one output per time point. The local feature window controls
    input context but not output granularity. We pair this formulation with a
    TSSB-derived benchmark in which development and test series are generated from
    disjoint UCR instances, so evaluation measures transfer to unseen instances
    rather than interpolation within a single graph.

\noindent \textbf{2. A controlled study of graph construction and aggregation.}
    We evaluate eleven time-series-to-graph transformations while holding the
    generated data, node features, GAT architecture, optimisation budget,
    initialisation policy, and seeds fixed. Matched GCN and same-feature MLP,
    random-forest, and 1-NN controls then examine the contributions of
    neighbour-specific aggregation and graph message passing.
    
\noindent \textbf{3. A practical operating regime for compact graph segmentation.}
     We show that the graph contributes context from beyond the local feature window rather than better features within it. At the downsampled benchmark resolution each segment is short and the fixed window already spans most of it, so WDPVG+GAT performs on par with the same-feature MLP, random forest, and 1-NN trained on the identical window. The graph helps only once a segment is longer than the window. The visibility graph then links each point to distant, non-adjacent samples in its line of sight, and can supply the context the window lacks. At native resolution, WDPVG+GAT varies less with window width than seq2point, outperforms the same-feature MLP at every tested window, and matches a window-searched seq2point at $70\times$ fewer parameters and $46\times$ less peak memory. 
     
These results position the graph model as a compact option for finely sampled data when model size and resolution-specific window tuning are important, rather than as a universal replacement for sequence models.

The paper is organised as follows. Section~\ref{sec:related_work} reviews related work. Section~\ref{sec:casting} elaborates on dense univariate segmentation, Section~\ref{sec:expsetup} describes the evaluation protocol, and Sections~\ref{sec:results} and~\ref{sec:discussion} present and interpret the results. Finally, Section~\ref{sec:conclusions} concludes the paper.


\section{Related Work}
\label{sec:related_work}

Time series segmentation (TSS) partitions a series into distinct, non-overlapping segments, each corresponding to an underlying state or pattern. Prior work falls into three categories~\cite{wang2024unsupervised}, the blocks of Table~\ref{tab:related_work}: Change Point Detection (CPD)~\cite{aminikhanghahi2017survey} locates change points, Windowed Segment Classification (WSC) labels fixed-length windows, and Dense Labeling or Precise Segmentation (DLPS) labels every time step. For each work, the table lists its task, data type, representation, model family, learning setting, and dataset count. A fourth block gathers learned models on visibility graphs, which do not segment but share our graph representation and so sit closest to our approach.

\begin{table*}[htbp]
\centering
\footnotesize
\setlength{\tabcolsep}{4pt}
\begin{tabular}{llllllc}
\toprule
\textbf{Work} & \textbf{Task} & \textbf{Data} & \textbf{Representation} & \textbf{Model} & \textbf{Supervision} & \textbf{\makecell{Datasets}} \\
\midrule
\multicolumn{7}{l}{\textit{Change point detection (CPD)}} \\
Killick et al. (2012)~\cite{killick2012optimal}      & CPD  & UTS & Raw signal       & Statistical (PELT)               & Unsupervised    & 1 \\
De Ryck et al. (2021)~\cite{de2021change}            & CPD  & UTS/MTS & Sliding window   & DL (Autoencoder)                 & Self-supervised & 7 \\
Ermshaus et al. (2023)~\cite{ermshaus2023clasp}      & CPD  & UTS & Sliding window   & ML (k-NN)                        & Self-supervised & 107 \\
Zheng et al. (2021)~\cite{zheng2021visibility}       & CPD  & UTS & Visibility graph   & Community detection              & Unsupervised    & 3 \\
Hu et al. (2023)~\cite{hu2023visibility}             & CPD  & MTS & Visibility graph   & Community detection (GA)         & Unsupervised    & 2 \\
\midrule
\multicolumn{7}{l}{\textit{Windowed segment classification (WSC)}} \\
Acharya et al. (2017)~\cite{acharya2017heartbeats}   & WSC  & UTS & Beat segment     & DL (CNN)                         & Supervised      & 1 \\
Phan et al. (2022)~\cite{phan2022sleeptransformer}   & WSC  & UTS & Spectrogram      & DL (Transformer)                 & Supervised      & 2 \\
Li et al. (2025)~\cite{li2025p2lhap}                 & WSC  & MTS & Patches          & DL (Transformer)                 & Supervised      & 3 \\
\midrule
\multicolumn{7}{l}{\textit{Dense labelling / precise segmentation (DLPS)}} \\
Yao et al. (2018)~\cite{yao2018efficient}            & DLPS & MTS & Raw signal       & DL (FCN)                         & Supervised      & 3 \\
Peimankar et al. (2021)~\cite{peimankar2021densecg}  & DLPS & UTS & Raw signal       & DL (CNN-BiLSTM)                  & Supervised      & 2 \\
Yoshimura et al. (2022)~\cite{yoshimura2022losnet}   & DLPS & MTS & Raw signal       & DL (TCN)                         & Supervised      & 1 \\
Gaugel et al. (2023)~\cite{gaugel2023prectime}       & DLPS & MTS & Sliding window   & DL (CNN-BiLSTM-CNN)                  & Supervised      & 1 \\
Goldbraikh et al. (2024)~\cite{goldbraikh2024mstcrnet} & DLPS & MTS & Raw signal     & DL (MS-TCN+RNN)                  & Supervised      & 3 \\
Shan et al. (2025)~\cite{shan2025utss}               & DLPS & UTS & Raw signal       & DL (U-Net)                      & Supervised      & 1 \\
\midrule
\multicolumn{7}{l}{\textit{Learned models on visibility graphs}} \\
Xuan et al. (2021)~\cite{xuan2021adaptive}           & TSC  & MTS & Visibility graph & DL (GNN)                        & Supervised      & 2 \\
Peng and Gao (2025)~\cite{peng2025vector}            & TSC  & MTS & Visibility graph & DL (GNN)                        & Supervised      & 1 \\
Chen et al. (2025)~\cite{chen2025mvgformer}          & General & MTS & Visibility graph & DL (Transformer)             & Supervised      & 25 \\
\midrule
\textbf{This Work} & \textbf{DLPS} & \textbf{UTS} & \textbf{Visibility graph} & \textbf{DL (GNN)} & \textbf{Supervised} & \textbf{59} \\
\bottomrule\end{tabular}
\caption{Summary of related work on time series segmentation, organised by task type (CPD/WSC/DLPS), together with learned models on visibility graphs. For each work we report the data type (univariate/multivariate), the underlying time-series representation, the model family, the learning setting (supervised / self-supervised / unsupervised), and the number of datasets used in its main evaluation. TSC denotes whole-signal time-series classification.}
\label{tab:related_work}
\end{table*}

Early CPD research relied on classical statistical methods that detect shifts in a parametric signal model, typically in its mean or variance. PELT~\cite{killick2012optimal}, the first of the five methods in the CPD block of Table~\ref{tab:related_work}, is a representative example: it recovers a globally optimal set of change points by minimising a penalised cost, unsupervised and directly on the raw signal. Like its peers, it requires a user-defined penalty and assumes the form of the change in advance. These limitations motivated the use of machine learning and deep learning methods. The following two rows list self-supervised methods: De Ryck et al.~\cite{de2021change} detect change points from increasing dissimilarity between windows of a learned time-invariant representation, and ClaSP~\cite{ermshaus2023clasp} recursively splits a series using a self-supervised classifier that requires no tuning. Its successor CLaP~\cite{ermshaus2025clap} extends this approach by also labelling the detected segments, merging those that a classifier repeatedly confuses, and a streaming variant, ClaSS~\cite{ermshaus2024class}, carries it to online data. The methods in the last two rows of the CPD block already build on graph representations of time series. Zheng et al.~\cite{zheng2021visibility} turn a univariate series into a visibility graph and identify its segments as communities, and Hu et al.~\cite{hu2023visibility} extend this idea to multivariate signals. Both, however, detect boundaries entirely through unsupervised community detection, without learning or semantic labelling.
 
To address CPD's inability to semantically label segments, later work recast TSS as a classification task. The WSC block in Table~\ref{tab:related_work} collects three supervised methods, in which the series is cut into fixed windows that are labelled one by one, as in heartbeat classification~\cite{acharya2017heartbeats}, sleep staging~\cite{phan2022sleeptransformer}, and human-activity recognition~\cite{li2025p2lhap}. In all of these approaches, temporal resolution is strictly bound by the window length, driving the need for finer-grained methods.
 
Shrinking the window to a single sample provides the finest granularity, DLPS, where every point is labelled. The DLPS block of Table~\ref{tab:related_work} gathers six such supervised segmenters. Their architectures are highly diverse, spanning fully-convolutional~\cite{yao2018efficient}, convolutional-recurrent~\cite{peimankar2021densecg, gaugel2023prectime, goldbraikh2024mstcrnet}, temporal-convolutional~\cite{yoshimura2022losnet}, and U-Net~\cite{shan2025utss} designs, yet all operate in the signal domain.
Signal-domain models learn dependencies over the temporal ordering of the series. A visibility graph supplies an additional, explicit topology in which shape-related samples can be connected even when they are far apart in time.

Graphs are already used widely in time-series modelling ~\cite{jin2024survey}, but most applications place variables or sensors on the nodes~\cite{wu2020connecting,cao2020stemgnn,khodayar2019windgraph,wang2020pm25gnn,jia2020graphsleepnet,baydilli2017hierarchical,baydilli2019world,munnix2012identifying}. Time then remains a separate axis. Our graph instead places the samples of one univariate series on the nodes, so its edges encode temporal shape relations directly.

Financial time-series studies provide a clear example of the variable-as-node formulation. Baydilli et al.~\cite{baydilli2017hierarchical} represented stocks in the BIST-100 market as nodes and used correlation-derived minimum-spanning and hierarchical trees to study market structure. Their later currency-network analysis~\cite{baydilli2019world} treated currencies as nodes and used their co-movement to examine relationships in the global foreign-exchange market. In these financial networks, nodes represent assets or currencies and edges encode statistical dependence. Our formulation differs by representing the time points of one univariate signal as nodes.

The closest segmentation studies also construct graphs over time points. Zheng et al.~\cite{zheng2021visibility} introduced WDPVG, and Hu et al.~\cite{hu2023visibility} extended visibility-graph segmentation to multivariate data. Both use unsupervised community detection rather than supervised node classification. Learned visibility-graph models have been studied for whole-series classification and other time-series tasks~\cite{xuan2021adaptive,peng2025vector,chen2025mvgformer}, but not for the dense univariate setting considered here. Temporal graph networks such as TGAT and TGN operate on interacting entities rather than on the graph of one signal~\cite{xu2020tgat,rossi2020tgn}.

We treat dense univariate segmentation as node classification on a visibility graph. WDPVG supplies a polarity-symmetric edge set, and GAT provides neighbour-specific aggregation over that fixed topology~\cite{zheng2021visibility,velickovic2018graph}. The task is supervised because each output is a semantic class label rather than only a boundary or an anonymous state. We evaluate the same model family on 59 datasets from several application domains. To our knowledge, no earlier study evaluates supervised point-wise labelling on the visibility graph of a univariate series.




\section{Dense Univariate Segmentation}
\label{sec:casting}

\subsection{Problem formulation}
\label{sec:prob:formulation}

Time series segmentation partitions a sequence into semantically meaningful regions. We formulate it as point-wise classification, also referred to as dense labelling, in which each time step receives a categorical segment label. This formulation maps directly to graph node classification~\cite{velickovic2018graph}: each time point is represented by one node, and segmentation becomes the assignment of labels to those nodes. Let $S$ be a univariate time series of length $N$:
\begin{equation}
    S = \{s_1, s_2, \dots, s_N\}, \quad s_i \in \mathbb{R}
\end{equation}
\noindent where each $s_i$ represents the value of the TS at time step $i$. The objective of TSS is to assign a label $y_i$ to each time point $s_i$, where the label corresponds to one of $C$ predefined segment classes:
\begin{equation}
    Y = \{y_1, y_2, \dots, y_N\}, \quad y_i \in \{0, 1, \dots, C-1\}
\end{equation}
We first construct a graph with the transformation $G=T(S)$. The graph $G=(V,E)$ represents the relations induced by the time series $S$, where:
\begin{itemize}
    \item $V = \{v_1, v_2, \dots, v_N\}$ is the set of nodes, where each node $v_i$ represents a time step $s_i$.
    \item  $E \subseteq V \times V$ is the set of edges that define the structural relationships between time points.
\end{itemize}

Given $G$, the objective is to assign a segment label $y_i$ to each node, with each node representing one time step. We define the segmentation model as a node-classification function $F_G$:
\begin{equation}
\label{eq:general_model}
    y_i = F_G(G; \theta)_i
\end{equation}
\noindent where $\theta$ represents the model parameters. This formulation aims to provide continuous and dense labelling across all time steps, enabling fine-grained segmentation.

\subsection{Time series to graph transformations}
\label{sec:transf}

The node-classification model $F_G$ of Eq.~\ref{eq:general_model} assigns a label to every node of the graph $G$, where each node is a single time step. We obtain this graph from a raw series in two steps. A transform $T$ produces the edges, and a standardised local window attached to each node provides its features. The edges and node features together form the graph that the GAT consumes. Both steps are collected in the BuildGraph routine of Algorithm~\ref{alg:graphrep}. The BuildGraph routine on lines~\ref{line:bg:build} to~\ref{line:bg:return} assembles the featured graph from two independent parts and returns them together. It applies the transform $T$ to obtain the edge set $\mathcal{E}$ together with the scalar edge attributes $\mathbf{e}$ used by the weighted transforms, and it calls WindowFeatures to obtain the node-feature matrix $X$.

\begin{algorithm}[t]
\caption{Graph construction and node features from a time series. Here $\mu_S$ and $\sigma_S$ are the mean and standard deviation of $S$, $\varepsilon$ is a small positive constant that guards against division by zero, and the edge attributes $\mathbf{e}$ are empty for unweighted transforms.}
\label{alg:graphrep}
\begin{algorithmic}[1]
\Function{BuildGraph}{$S,T,w$}
  \State $(\mathcal{E},\mathbf{e})\gets T(S)$ \Comment{edges and attributes from the transform} \label{line:bg:build}
  \State $X\gets\Call{WindowFeatures}{S,w}$ \Comment{node features} \label{line:bg:feat}
  \State \Return $(\mathcal{E},\mathbf{e},X)$ \Comment{featured graph: edges, attributes, node features} \label{line:bg:return}
\EndFunction
\Statex
\Function{WindowFeatures}{$S,w  $}
  \State $z_i\gets(s_i-\mu_S)/(\sigma_S+\varepsilon)\ \forall i$ \Comment{per-series $z$-standardisation} \label{line:wf:z}
  \State $\tilde z\gets\Call{EdgePad}{z,w}$ \Comment{edge-pad out-of-range positions}
  \State \Return $\big[\,x_i=(\tilde z_{i-w},\dots,\tilde z_{i+w})\,\big]_{i=1}^{N}\in\mathbb{R}^{N\times(2w+1)}$ \Comment{local window per node} \label{line:wf:return}
\EndFunction
\end{algorithmic}
\end{algorithm}

The WindowFeatures routine on lines~\ref{line:wf:z} to~\ref{line:wf:return} builds these node features. A single scalar per node carries little information, so each node instead receives a standardised \emph{local window} of the raw series. WindowFeatures $z$-standardises the series once and edge-pads the out-of-range positions, so that even boundary nodes have a full window. Node $i$ then takes the $2w+1$ padded standardised values centred on it, $x_i=(\tilde z_{i-w},\dots,\tilde z_i,\dots,\tilde z_{i+w})$. These features give the GAT direct access to the local waveform around each point while remaining strictly label-free. The transform $T$ and the half-width $w$ are the only inputs that vary across the representations we study. The remainder of the construction is identical for every transform. The transform $T$ is therefore the single component that distinguishes the graph representations. Following Silva et al.~\cite{silva2021time}, we consider three families of such transforms, each producing the edges of $G$ from $S$ in a different way. Their formal definitions and a summary table are given in Appendix~A.

Visibility graphs connect points by line of sight. Each sample is represented at its time coordinate, and two samples are linked when the straight line between them lies above every intervening sample~\cite{lacasa2008time}. The Weighted Dual-Perspective Visibility Graph (WDPVG) of Zheng et al.~\cite{zheng2021visibility} combines the natural visibility graph of the original series with that of the reflected series. The union represents peaks and troughs symmetrically. In our implementation, each retained edge is weighted by the Euclidean distance between its endpoints over the time and value axes. We combine the two weighted adjacency matrices by their elementwise maximum and normalise the retained weights within each graph by the maximum edge weight. The dual perspective changes the edge set. An edge present in both perspectives keeps the same distance weight. Horizontal~\cite{luque2009horizontal} and directed~\cite{lacasa2012time} variants are evaluated with the same model.

Transition-based mappings take a more indirect view while retaining one node per time point. The signal is discretised into states, for example by quantile binning or ordinal patterns, and a state-transition matrix is estimated. This matrix is then lifted back to the time axis as a Markov transition field: the edge value between time-point nodes $i$ and $j$ is the estimated transition probability between the states assigned to those two points. The resulting graph therefore has $N$ time-indexed nodes and supports a separate prediction at every point, although its edges encode state-transition dynamics rather than direct geometric visibility.

Proximity networks retain the same time-point node set, embed each point in a reconstructed phase space, and connect time points whose embedded vectors are close under a distance or similarity measure, as in the k-nearest-neighbour network~\cite{donner2011recurrence}. These connections can represent recurring regimes, although distance-based neighbourhoods may smooth local boundaries. Among the three families, only the visibility constructions require no dataset-specific discretisation level, neighbourhood radius, or embedding dimension.

\subsection{Proposed segmentation model}
\label{sec:model}

The graph transform maps the time series to $G$, with one node for each time point. The segmentation model assigns one class label to every node. The GAT is the learned, transform-agnostic component and is kept unchanged across the graph representations described in Section~\ref{sec:transf}.
Formally, the prediction function $F_G$ maps $G$ to the node-label vector $Y_{1 \times N}$. To realise $F_G$ in Eq.~\ref{eq:general_model}, we use a Graph Attention Network (GAT)~\cite{velickovic2018graph}. Its attention coefficients provide learned, neighbour-specific aggregation over the fixed graph topology. The controlled GAT--GCN comparison in Section~\ref{sec:train_eval} tests whether this architecture improves performance in the present task.

\begin{figure}[htbp]
    \centering
    \includegraphics[width=0.4\linewidth]{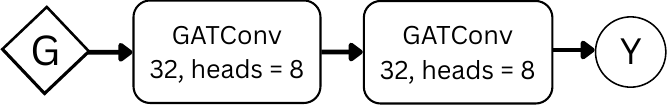}
    \caption{Selected GAT architecture for point-wise time-series segmentation.}
    \label{fig:model:gnn}
\end{figure}

As shown in Figure~\ref{fig:model:gnn}, the selected architecture contains two hidden GAT layers of width $32$ with eight concatenated heads, followed by a single-head output layer that maps each node to $C$ class logits. Throughout the manuscript, network depth denotes the number of hidden message-passing layers and excludes the output layer. The hidden layers use ELU activations. Weighted graph variants supply their normalised scalar edge weights to every GAT layer, whereas unweighted variants provide only the graph topology. The final logits are used directly in the cross-entropy loss. Softmax is applied only when class probabilities are required. This configuration was selected as the Pareto-front knee of the validation-only architecture search described in Section~\ref{sec:train_eval}. At approximately $89.6$k parameters, it is about $70\times$ smaller than seq2point.

\subsection{Selected baseline models}
\label{sec:baseline}

Many deep segmentation models operate on sliding windows. As the closest supervised pointwise window reference, we use a multiclass adaptation of the seq2point architecture of Zhang et al.~\cite{zhang2018sequence}. We retain its published five-layer convolutional configuration: 30, 30, 40, 50, and 50 filters with kernel widths 10, 8, 6, 5, and 5, followed by a 1024-unit dense layer. The original model performs scalar midpoint regression, so we replace its scalar output with a linear layer that produces $C$ class logits. The convolutions in our implementation are valid. Replicate padding and stride-one windowing yield one prediction for every original time point. Seq2point is trained with Adam and plain cross-entropy in batches of 512 for at most 300 epochs. Its window length and learning rate are selected on the fitting and validation partitions using the development protocol described in Section~\ref{sec:train_eval}. This validation-based hyperparameter search is distinct from the native-resolution window-sensitivity analysis in Section~\ref{sec:resolution}, where the models are evaluated over a common range of receptive-window widths. After the final seq2point settings are fixed, each training run reserves 20\% of the generated training series for early stopping on weighted F1. This holdout is used only to select the stopping checkpoint, with patience 40 after a minimum of 50 epochs, and is not used for architecture or hyperparameter selection.

To isolate the contribution of graph message passing, we additionally evaluate three non-graph controls on the identical per-point, per-series $z$-standardised 81-dimensional window features used by the selected GAT. The MLP has hidden layers $(256,128)$ with ReLU activations and is trained with Adam (learning rate $10^{-3}$, $L_2$ penalty $10^{-4}$) for at most 300 iterations. The random forest uses 200 unpruned trees, $\sqrt{d}$ candidate features per split, bootstrap sampling, and balanced class weights. The 1-NN classifier uses uniform Euclidean distance over the same standardised feature vectors. The matched GCN has the same two-hidden-layer depth and 256-dimensional hidden representation as the GAT, uses symmetric graph normalisation with self-loops, and receives the same WDPVG edge weights. It is trained with the same optimiser, learning rate, weight decay, epoch budget, and class-weighted loss as the GAT, so the GAT--GCN comparison changes the aggregation architecture while holding the available graph information fixed.

As an unsupervised reference, we use a two-stage pipeline denoted \emph{ClaSP+KM}. ClaSP~\cite{ermshaus2023clasp} first detects change points and partitions the series. We then represent each detected segment by its mean, standard deviation, minimum, maximum, and linear slope, and cluster these descriptors with $k$-means. The resulting cluster identifier is assigned to every time point in the segment. We set $k$ to the ground-truth number of classes so that the comparison evaluates boundary detection and state grouping rather than selection of the number of states. This constructed reference should not be confused with the published CLaP method~\cite{ermshaus2025clap}, which is discussed separately in Section~\ref{sec:related_work}.

We also report classical unsupervised references: PELT, Binary Segmentation, and Window-based change-point detection~\cite{truong2020ruptures}, together with FLUSS matrix-profile regime detection~\cite{gharghabi2017fluss}. Binary Segmentation, the Window method, and FLUSS are supplied the ground-truth number of segments. PELT and ClaSP infer their own change points. The supervised models (GAT, seq2point, MLP, random forest, and 1-NN) predict the shared semantic class indices directly and are scored without relabelling. Only the unsupervised methods produce arbitrary segment or cluster identifiers. For each test series separately, these identifiers are aligned to the ground-truth labels by Hungarian assignment on the label-contingency matrix, choosing the assignment that maximises agreement, and the aligned per-point predictions are then scored. This is an oracle label-permutation step used only for evaluation. The resulting per-series scores are averaged. Alignment is never performed jointly across test series. These methods are contextual references rather than task-matched supervised competitors.

We did not identify a published method that matches all parts of the present evaluation: supervised semantic labelling, one output per time point, univariate input, and testing on unseen instances generated with the TSSB procedure. We do not treat the comparison as an external state-of-the-art ranking. Seq2point is the closest supervised window-based dense-labelling baseline. The MLP, random forest, and 1-NN controls use the same point features and test the contribution of graph message passing. The matched GCN tests the choice of graph aggregation. The unsupervised methods provide contextual reference points for change-point and state-detection settings. All supervised methods follow the same inductive data protocol.

\section{Methods}
\label{sec:expsetup}
This section describes dataset construction, model development, and final evaluation.

\subsection{Dataset}
\label{sec:dataset}

We evaluated the method on a TSSB-derived inductive benchmark constructed from the same UCR sources and per-dataset generation specifications as TSSB~\cite{ermshaus2023clasp}. The original TSSB collection contains 75 annotated univariate series with between 1 and 9 segments. We excluded series containing only one segment because they have no change point and do not test segmentation performance. 
We also excluded series with duplicated segment classes because the distributed TSSB annotations provide change-point locations but not semantic labels that distinguish repeated occurrences. The resulting evaluation contains 59 TSSB-derived datasets spanning electrocardiography and other physiological signals, gesture and motion capture, appliance and robot sensors, image-outline profiles, and food or material spectra. By segment count, the benchmark contains 21 two-segment datasets, 13 three-segment datasets, 9 four-segment datasets, 10 five-segment datasets, and 6 datasets with six or seven segments. This breadth evaluates the same model family across domains, but does not constitute cross-domain transfer because a separate model is trained for each dataset.

For the proposed method, each generated series was transformed into the graph representations described in Section~\ref{sec:transf}. At the benchmark's default downsampled resolution, seq2point used the validation-selected window length of 151 samples. The input series was replicate-padded by 75 samples on both sides (the boundary value was repeated), the window was advanced with stride 1, and the centre point of every padded window was classified. Seq2point and the graph-based methods were therefore evaluated on the same complete set of time points, including the original first and last samples. 

\begin{algorithm}[htbp]
\caption{TSSB-derived series generation.}
\label{alg:genseries}
\begin{algorithmic}[1]
\Function{GenSeries}{$\mathrm{Split},\mathcal{C},N,\mathrm{seed}$}
  \State initialise an unused-instance pool $\mathcal{U}_c\gets\mathrm{Split}_c$ for every $c\in\mathcal{C}$ \label{line:gs:pool}
  \For{$n\gets 1$ \textbf{to} $N$}
    \State draw one instance from each $\mathcal{U}_c$ and remove it from that pool \label{line:gs:draw}
    \State resample each selected instance according to the TSSB configuration
    \State concatenate the resulting segments in the TSSB-prescribed order $\Rightarrow(S_n,Y_n)$ \label{line:gs:concat}
  \EndFor
  \State \Return $\{(S_n,Y_n)\}_{n=1}^{N}$ \Comment{sampling without replacement across the generated collection}
\EndFunction
\end{algorithmic}
\end{algorithm}

\noindent\textbf{TSSB-derived inductive series generation.} TSSB~\cite{ermshaus2023clasp} constructs each benchmark series from a source dataset in the UCR archive~\cite{dau2019ucr} by selecting a specified class subset, drawing one labelled instance for each prescribed segment, resampling it according to the TSSB configuration, and concatenating the resulting subsequences in a fixed, dataset-specific order. Concatenation points are the ground-truth boundaries, and every time point inherits the class label of its source instance. Algorithm ~\ref{alg:genseries} presents the generation we use. We keep the same class subsets, resampling factors, segment counts, and fixed segment orders, but repeat the procedure to obtain multiple generated series. The draw step on line ~\ref{line:gs:draw} removes each selected instance from an unused-instance pool, so within each data partition and repetition the class-specific source instances are sampled without replacement across the complete generated collection, and the concatenation step on line~\ref{line:gs:concat} joins them in the prescribed order. Each selected instance contributes one segment to one generated series only. Consequently, no raw instance or generated segment is repeated within the fitting, validation, or test collection for that repetition. We refer to the resulting collection as a \emph{TSSB-derived inductive benchmark}, rather than the original set of single distributed TSSB series. The official UCR \textsc{Train} split is used for fitting and validation, whereas test series are generated exclusively from the disjoint official UCR \textsc{Test} instances.

\subsection{Model training and evaluation}
\label{sec:train_eval}

We evaluate every method under an \emph{inductive} protocol. The official UCR \textsc{Train} instances are used only for model development and are partitioned into disjoint fitting and validation pools. The official UCR \textsc{Test} instances remain untouched during architecture and hyperparameter selection. Training, validation, and test series are converted into separate graphs that share no nodes or edges. This prevents message passing across data partitions and evaluates generalisation to unseen raw instances rather than interpolation between labelled nodes of one graph.

Algorithm~\ref{alg:pipeline} gives the full inductive pipeline. Line~\ref{line:select} selects the architecture and hyperparameters on the validation split alone, with no access to the test instances. The remaining lines repeat an evaluation loop over the five seeds. Within each seed, GenSeries produces disjoint training and test series, BuildGraph converts every series into its featured graph, and the selected GAT is trained on the training graphs, on lines~\ref{line:gen} to ~\ref{line:train}. Every held-out node is then labelled and each test series scored with weighted F1, on lines~\ref{line:predict} to ~\ref{line:score}. The pipeline returns the mean and standard deviation of the per-seed scores across the five seeds on line~\ref{line:return}.

\begin{algorithm}[t]
\caption{Inductive training and evaluation pipeline.}
\label{alg:pipeline}
\begin{algorithmic}[1]
\Require UCR \textsc{Train}/\textsc{Test} split; class subset $\mathcal{C}$; transform $T$; candidate configs $\mathcal{H}$; seeds $\mathcal{S}$; sizes $N_{\mathrm{tr}},N_{\mathrm{te}}$; epochs $E$.
\Ensure Mean and standard deviation of per-series weighted F1 on \textsc{Test}.
\State $h^{*}\gets\Call{SelectOnValidation}{\textsc{Train},0.7,0.3,\mathcal{C},T,\mathcal{H},\mathcal{S}}$ \Comment{validation-only grid search, no \textsc{Test} access} \label{line:select}
\State $\mathrm{seedF1}\gets[\,]$
\For{$\mathrm{seed}\in\mathcal{S}$}
  \State $\mathcal{D}_{\mathrm{tr}}\gets\Call{GenSeries}{\textsc{Train},\mathcal{C},N_{\mathrm{tr}},\mathrm{seed}}$;\ \ $\mathcal{D}_{\mathrm{te}}\gets\Call{GenSeries}{\textsc{Test},\mathcal{C},N_{\mathrm{te}},\mathrm{seed}}$ \Comment{disjoint instances, Alg.~\ref{alg:genseries}} \label{line:gen}
  \State $G_{\bullet}\gets\{(\Call{BuildGraph}{S,T,h^{*}.w},Y):(S,Y)\in\mathcal{D}_{\bullet}\}$ for $\bullet\in\{\mathrm{tr},\mathrm{te}\}$ \Comment{featured graphs, Alg.~\ref{alg:graphrep}} \label{line:graphs}
  \State $g_{\theta}\gets\Call{TrainGAT}{G_{\mathrm{tr}},h^{*},E}$ \Comment{selected GAT, class-weighted CE, Adam} \label{line:train}
  \State $f\gets[\,]$
  \ForAll{$(G,Y)\in G_{\mathrm{te}}$}
     \State $\hat Y\gets\arg\max_{c} g_{\theta}(G)$ \Comment{one prediction per held-out node} \label{line:predict}
     \State $f.\mathrm{append}(\Call{WeightedF1}{Y,\hat Y})$ \label{line:score}
  \EndFor
  \State $\mathrm{seedF1}.\mathrm{append}(\mathrm{mean}(f))$
\EndFor
\State \Return $\mathrm{mean}(\mathrm{seedF1}),\ \mathrm{std}(\mathrm{seedF1})$ \label{line:return}
\end{algorithmic}
\end{algorithm}

\noindent\textbf{Sample size, validation, and repeated evaluation.} For architecture and hyperparameter selection, the official UCR \textsc{Train} instances are divided at the raw-instance level into a stratified 70:30 fitting/validation split. This selection procedure is repeated with five seeds $\{42,5,820,7,101\}$, and configurations are ranked by their mean validation weighted F1 across the five repetitions. The official UCR \textsc{Test} instances are not accessed during selection. After the final architecture and hyperparameters are chosen, the graph models are refitted on series generated from the complete official UCR \textsc{Train} split and evaluated on series generated exclusively from the official UCR \textsc{Test} split. Each generated series contains one selected instance of every included class, with one segment formed from each instance and the segments arranged in the fixed order prescribed by TSSB. Source instances are sampled without replacement across the generated collection, so the number of series is bounded by the least populated included class. We form $N_{\mathrm{test}}=\min(\min_{c}|\textsc{Test}_c|,\,40)$ test series. Preserving the fixed order makes the generation procedure faithful to TSSB, and no model receives an explicit absolute-position feature. The evaluation nevertheless measures unseen-instance generalisation only under the benchmark-prescribed order. It does not test unseen segment permutations. Each generated series is transformed into its own graph and labelled per point. The five final repetitions redraw the generated series and model initialisation, so they estimate generation and optimisation variability rather than constituting disjoint cross-validation folds.

\noindent\textbf{Model selection, training, and testing.} Architecture and hyperparameter selection use only the stratified 70:30 fitting and validation graphs generated from the official UCR \textsc{Train} instances. Validation performance is averaged across five seeds. After selection, the GAT configuration is frozen and refitted on series generated from the complete official UCR \textsc{Train} pool. It is then evaluated without further tuning on the $N_{\mathrm{test}}$ graphs generated from the official UCR \textsc{Test} instances. Each graph is processed in full. The GAT and matched GCN use Adam for 300 fixed epochs, a learning rate of $5\times10^{-4}$, weight decay of $5\times10^{-4}$, and inverse-frequency weighted cross-entropy. Weighted visibility variants supply a scalar edge channel to all graph layers. Unweighted variants use topology alone. Seq2point hyperparameters are selected with the same development split. In each final seq2point run, 20\% of the generated training series are reserved only for early stopping and checkpoint selection. The software environment, seeds, hardware, and public code release are described in Section~\ref{sec:reproducibility}.

For the GAT, the grid covers hidden-layer depth $\{2,4,6\}$, hidden width $\{32,96,192\}$, and attention heads $\{2,4,8\}$. We test the role of attention separately by replacing the selected GAT with a GCN of matched depth and hidden dimension. At the default downsampled resolution, the node-feature half-width is evaluated over $w\in\{0,5,10,20,40\}$ at a fixed reference architecture. The GAT learning rate and weight decay remain fixed at $5\times10^{-4}$.

For the main benchmark, seq2point hyperparameters are selected over window lengths $W\in\{35,51,101,151\}$ and learning rates $\{10^{-3},3\times10^{-4}\}$ using the same fitting and validation partitions. The selected values are $W=151$ and a learning rate of $3\times10^{-4}$. The 20\% holdout used for seq2point early stopping is introduced only after these settings are fixed, during each final training run. It does not contribute to architecture or hyperparameter selection. Separately, the native-resolution window-sensitivity analysis evaluates WDPVG+GAT, seq2point, and the MLP over the broader common grid $W\in\{35,51,101,151,201,275\}$. This sweep measures robustness to local-context width and is not used to select the main downsampled configuration. All reported test scores use configurations chosen without access to the test instances.

\noindent\textbf{Comparing graph representations.} We compare the visibility, transition, and proximity transforms of Section~\ref{sec:transf} as competing inputs to one fixed segmentation architecture. The graph transform is the only experimental factor. All variants use the same fitting, validation, and test series, node features, parameter-initialisation policy, GAT depth, width, and head count, optimiser, training budget, seeds, and metric. Architecture selection was performed on WDPVG using fitting and validation data only, after which the selected architecture was held fixed for all eleven graph transformations.

\noindent\textbf{Evaluation metric and statistical analysis.} As the task is dense per-point classification over imbalanced, multi-class segments, we report support-weighted F1. Each method is scored independently on every generated test series. The scores are then averaged over test series and over the five seeds. Each observation entering an inferential test is therefore one dataset's weighted F1 averaged over the five seeds, giving $n=59$ paired observations. Planned pairwise comparisons use two-sided Wilcoxon signed-rank tests at $\alpha=0.05$, with zero paired differences discarded before ranking. Effect sizes are matched-pairs rank-biserial correlations, $r=(W_+-W_-)/(W_++W_-)$. Confidence intervals on mean gaps are 95\% percentile bootstrap intervals from 10,000 resamples of the 59 paired per-dataset differences. Family-wise error is controlled with Holm correction within six prespecified families. These comprise the five main supervised contrasts against WDPVG+GAT, the five contextual unsupervised contrasts against WDPVG+GAT, the six post-hoc contrasts among the four natural-visibility variants after the omnibus test, the six GAT--seq2point window-size contrasts, the six parallel GAT--MLP window-size contrasts, and the four resolution contrasts. The Wilcoxon test is used because the observations are paired by dataset and no normal distribution of the paired gaps is assumed. Transform comparisons use Friedman repeated-measures tests followed, where appropriate, by Holm-adjusted paired post-hoc tests.

\subsection{Reproducibility}
\label{sec:reproducibility}
The source code, experiment configurations, and software requirements are publicly available at \url{https://github.com/sensorlab/TS-GRAPH-SEGMENTATION}. The graph models were implemented with PyTorch and PyTorch Geometric, and the baseline controls with scikit-learn. Training and evaluation were conducted on a workstation equipped with an Intel Xeon Gold CPU and an NVIDIA A100 GPU (80GB VRAM). The GAT models were trained using five fixed random seeds $\{42,5,820,7,101\}$, as described in Section~\ref{sec:expsetup}.

\begin{table*}[!htbp]
\centering
\caption{Weighted F1 of the eleven time-series-to-graph transformations under one fixed $89.6$k-parameter GAT, stratified by the number of segments (datasets per tier in parentheses). Only the graph transform changes: every entry uses the same generated fitting, validation, and test series, node features, parameter-initialisation policy, architecture, optimiser, training budget, and five seeds. ``All'' is the overall mean $\pm$ standard deviation over seeds.}
\label{tab:transforms}
{
\begin{tabular}{lrcccccc}
\toprule
Method & \#Params & 2 (21) & 3 (13) & 4 (9) & 5 (10) & 6--7 (6) & All \\
\midrule
NVG (directed)        & 89.6k & 0.932 & 0.890 & 0.933 & 0.914 & 0.904 & 0.917 $\pm$ 0.002 \\
\textbf{WDPVG}        & 89.6k & 0.920 & 0.914 & 0.940 & 0.885 & 0.927 & 0.916 $\pm$ 0.006 \\
NVG (weighted)        & 89.6k & 0.929 & 0.893 & 0.936 & 0.910 & 0.906 & 0.916 $\pm$ 0.002 \\
NVG                   & 89.6k & 0.917 & 0.910 & 0.940 & 0.876 & 0.909 & 0.911 $\pm$ 0.002 \\
HVG (weighted)        & 89.6k & 0.917 & 0.867 & 0.942 & 0.920 & 0.896 & 0.908 $\pm$ 0.001 \\
HVG (directed)        & 89.6k & 0.918 & 0.866 & 0.943 & 0.919 & 0.898 & 0.908 $\pm$ 0.002 \\
HVG                   & 89.6k & 0.913 & 0.878 & 0.932 & 0.898 & 0.900 & 0.904 $\pm$ 0.003 \\ \hline
Quantile              & 89.6k & 0.907 & 0.905 & 0.956 & 0.648 & 0.934 & 0.873 $\pm$ 0.003 \\ 
OPTN                  & 89.6k & 0.909 & 0.899 & 0.957 & 0.646 & 0.936 & 0.872 $\pm$ 0.003 \\
CGPS                  & 89.6k & 0.906 & 0.898 & 0.957 & 0.649 & 0.932 & 0.871 $\pm$ 0.002 \\ \hline
Proximity                & 89.6k & 0.739 & 0.613 & 0.522 & 0.405 & 0.314 & 0.578 $\pm$ 0.009 \\
\bottomrule
\end{tabular}
}
\end{table*}
\begin{figure}[t]
\centering
\includegraphics[width=0.6\linewidth]{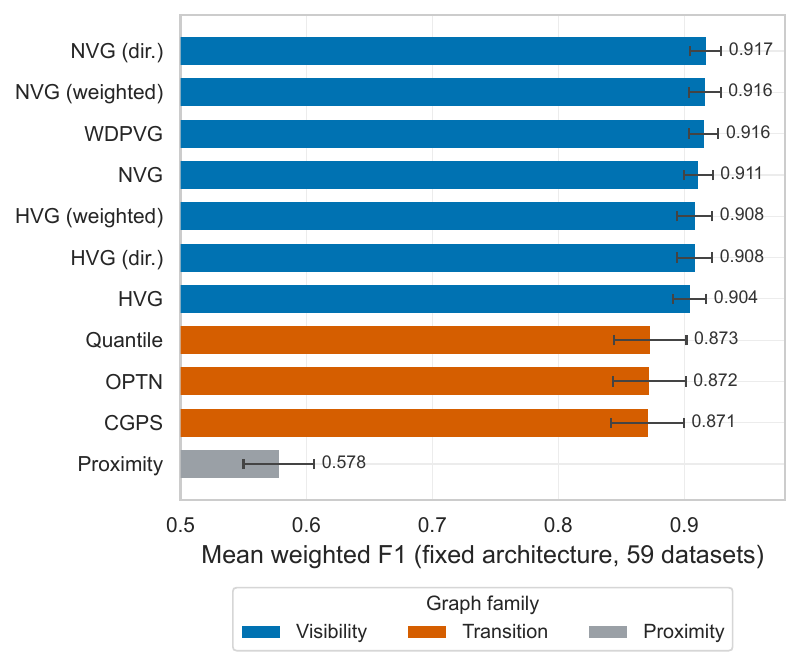}
\caption{Comparison of the eleven TS-to-graph transformations under one fixed GAT architecture (mean weighted F1 over the 59 datasets, error bars one standard error across datasets). Under the evaluated graph-construction settings, the visibility variants have higher mean F1 than the transition and proximity transforms. Within the natural-visibility variants, WDPVG, directed NVG, and weighted NVG form a statistically indistinguishable top group after Holm correction. WDPVG is retained as a polarity-symmetric representative rather than as a uniquely best or most efficient transform.}
\label{fig:transform}
\end{figure}

\section{Results}
\label{sec:results}

This section evaluates the network-science formulation of dense univariate segmentation introduced in Section~\ref{sec:casting} under the protocol described in Section~\ref{sec:expsetup}. Section~\ref{sec:selecting} examines sensitivity to the three graph-transformation families introduced in Section~\ref{sec:transf}. Section~\ref{sec:overall} then evaluates the WDPVG+GAT model from Section~\ref{sec:model} against the supervised controls, matched GCN, and contextual unsupervised references defined in Section~\ref{sec:baseline}, across all 59 datasets described in Section~\ref{sec:dataset}. Section~\ref{sec:resolution} examines window-width robustness, temporal resolution, and computational cost. Section~\ref{sec:results_ablation} reports the remaining component ablations and architecture search.

\subsection{Sensitivity to the graph representation}
\label{sec:selecting}

The graph representation is a principal design choice. Preliminary experiments restricted to the fitting and validation partitions indicated that WDPVG was comparatively stable across local-context settings, so it was fixed as the representative transform for the detailed benchmark before final test evaluation. We nevertheless assess sensitivity to that choice by holding the architecture fixed and varying only the graph transform over all eleven candidates. Table~\ref{tab:transforms} and Figure~\ref{fig:transform} report the resulting test performance. This analysis measures sensitivity to the transform choice and is not used to select the reported model retrospectively.

\begin{table*}[!htbp]
\centering
\caption{Per-transformation computational cost under the fixed
$89.6$k-parameter GAT, averaged over the 59 datasets and five seeds and listed in the order of
Table~\ref{tab:transforms}. All rows share the same model, so differences arise from the graph each
transform produces. \emph{Edges} counts \texttt{edge\_index} entries per test graph (directed
transforms store each relation once, whereas undirected transforms store both orientations), and
therefore reflects message-passing cost. \emph{Build} is graph construction for the test split,
\emph{Train} is the 300-epoch GAT training time, \emph{Infer} is inference over the test split, and
\emph{Peak mem} is peak GPU memory measured with \texttt{torch.cuda.max\_memory\_allocated}. The
wall-clock values come from one common all-transform sweep and should be compared within this table.
The independent main benchmark profile in Table~\ref{tab:efficiency} records 83.2~s for WDPVG rather
than 67~s. Stored-edge count and peak memory agree across the two runs, while wall-clock time varies with the execution environment and system load.}
\label{tab:transform_efficiency}
{
\setlength{\tabcolsep}{5pt}
\begin{tabular}{lrrrrrr}
\toprule
Transform & F1 & Edges & Build (ms) & Train (s) & Infer (ms) & Peak mem (MB) \\
\midrule
NVG (directed)         & $0.917$ & $1{,}873$ & $70$ & $57$ & $93$ & $31.7$ \\
\textbf{WDPVG}         & $0.916$ & $6{,}348$ & $132$ & $67$ & $103$ & $73.1$ \\
NVG (weighted)         & $0.916$ & $1{,}873$ & $87$ & $67$ & $104$ & $36.6$ \\
NVG                    & $0.911$ & $3{,}745$ & $51$ & $52$ & $82$ & $43.2$ \\
HVG (weighted)         & $0.908$ & $380$ & $74$ & $70$ & $111$ & $23.7$ \\
HVG (directed)         & $0.908$ & $380$ & $31$ & $53$ & $82$ & $22.4$ \\
HVG                    & $0.904$ & $761$ & $41$ & $58$ & $93$ & $24.5$ \\
\midrule
Quantile               & $0.873$ & $7{,}924$ & $53$ & $76$ & $119$ & $80.3$ \\
OPTN                   & $0.872$ & $7{,}924$ & $61$ & $66$ & $102$ & $80.3$ \\
CGPS                   & $0.871$ & $7{,}924$ & $58$ & $73$ & $113$ & $80.3$ \\
\midrule
Proximity              & $0.578$ & $2{,}086$ & $90$ & $73$ & $114$ & $36.4$ \\
\bottomrule
\end{tabular}
}
\end{table*}

Under the evaluated graph-construction settings, the visibility variants occupy the highest mean ranks, followed by the transition networks (quantile, OPTN, and CGPS, at ${\sim}0.87$ mean weighted F1) and the proximity ($k$-NN) graph ($0.58$). The eleven-transform Friedman test rejects equality among the transforms ($p=2\times10^{-33}$), but it is not itself a direct test of a visibility-versus-transition family contrast. The seven visibility variants are not all equivalent (Friedman $\chi^2=22.3$, $p=0.0011$), and the result remains significant within the four natural-visibility variants ($\chi^2=9.6$, $p=0.022$). However, Holm-adjusted pairwise tests identify a top cluster: WDPVG is indistinguishable from directed NVG ($p_{\mathrm{Holm}}=1.0$, $r=-0.06$) and weighted NVG ($p_{\mathrm{Holm}}=1.0$, $r=-0.09$). Its raw comparison with plain NVG ($p_{\mathrm{raw}}=0.040$) does not survive correction ($p_{\mathrm{Holm}}=0.23$). Plain NVG and the horizontal-visibility variants have the lower ranks, while the three leading transforms are not separated by the corrected pairwise tests. We retain WDPVG for the detailed experiments because its dual perspective represents peaks and troughs symmetrically and requires no dataset-specific graph hyperparameter. It is not selected on the basis of superior accuracy or computational efficiency.

The cost measurements in Table~\ref{tab:transform_efficiency} show that density, rather than model size, governs graph memory. Directed NVG is the most attractive accuracy--efficiency point among the leading transforms ($0.917$ F1, $1{,}873$ stored edges, and $31.7$~MB), whereas WDPVG is the heaviest visibility variant ($6{,}348$ stored edges and $73.1$~MB) for essentially the same F1. The horizontal-visibility graphs are the sparsest and lightest, at a small accuracy cost, while the transition graphs are both the densest and less accurate. All transforms build in under $0.15$~s and train in approximately one minute in the common sweep. The 67~s WDPVG value is retained for within-sweep comparability. An independent main-profile run reported in Table~\ref{tab:efficiency} takes 83.2~s while producing the same stored-edge count and peak-memory value. WDPVG is retained for its polarity-symmetric construction, not for computational efficiency.

\subsection{Benchmarking the proposed model: visibility-graph GAT}
\label{sec:overall}

\begin{table*}[!htbp]
\centering
\caption{Main benchmark: weighted F1 (mean over the 59 datasets, stratified by the number of segments, with datasets per tier in parentheses) of the representative visibility-graph GAT (VG+GAT with the WDPVG transform) against the attention-free GCN ablation, supervised non-graph controls, and an unsupervised contextual panel. Some unsupervised methods are supplied the segment count, whereas PELT and ClaSP infer change points. ``All'' is the overall mean $\pm$ standard deviation over the five seeds. The statistical column reports the mean paired difference WDPVG+GAT minus the row method, followed by its 95\% bootstrap confidence interval. Positive values favour WDPVG+GAT. The final column gives two-sided Wilcoxon signed-rank $p$-values with Holm correction. The five supervised and five unsupervised comparisons form separate correction families because they address different learning settings. Bold $p$-values remain significant at $\alpha=0.05$. Table~\ref{tab:transforms} compares the eleven graph transforms, and Section~\ref{sec:resolution} examines resolution dependence.}
\label{tab:main_results}
{
\footnotesize
\setlength{\tabcolsep}{2.8pt}
\begin{tabular}{lrcccccccc}
\toprule
Method & \#Params & 2 (21) & 3 (13) & 4 (9) & 5 (10) & 6--7 (6) & All & \shortstack{GAT $-$ method\\(95\% CI)} & $p_{\mathrm{Holm}}$ \\
\midrule
\multicolumn{10}{l}{\emph{Supervised: visibility-graph GAT (ours)}}\\
\textbf{WDPVG+GAT} & 89.6k & 0.920 & 0.914 & 0.940 & 0.885 & 0.927 & 0.916 $\pm$ 0.006 & Reference & --- \\
\midrule
\multicolumn{10}{l}{\emph{Supervised: graph $+$ GCN (attention ablation)}}\\
WDPVG+GCN & 87.6k & 0.889 & 0.887 & 0.910 & 0.805 & 0.888 & 0.877 $\pm$ 0.002 & $+0.039\ [0.029,\ 0.049]$ & $\mathbf{9.0\times10^{-9}}$ \\
\midrule
\multicolumn{10}{l}{\emph{Supervised: non-graph baselines}}\\
seq2point & 6.29M & 0.955 & 0.923 & 0.972 & 0.956 & 0.957 & \textbf{0.951} $\pm$ 0.002 & $-0.034\ [-0.048,\ -0.020]$ & $\mathbf{2.3\times10^{-6}}$ \\
1-NN & --- & 0.920 & 0.870 & 0.944 & 0.934 & 0.930 & 0.916 $\pm$ 0.002 & $+0.0002\ [-0.014,\ 0.015]$ & $1.00$ \\
MLP & 54.3k & 0.918 & 0.862 & 0.946 & 0.939 & 0.915 & 0.913 $\pm$ 0.002 & $+0.003\ [-0.014,\ 0.027]$ & $1.00$ \\
Random forest & --- & 0.908 & 0.860 & 0.925 & 0.918 & 0.934 & 0.904 $\pm$ 0.002 & $+0.012\ [-0.003,\ 0.029]$ & $1.00$ \\
\midrule
\multicolumn{10}{l}{\emph{Unsupervised contextual references}}\\
Window & --- & 0.759 & 0.690 & 0.706 & 0.503 & 0.648 & 0.681 $\pm$ 0.002 & $+0.235\ [0.198,\ 0.275]$ & $\mathbf{1.2\times10^{-10}}$ \\
FLUSS & --- & 0.567 & 0.481 & 0.467 & 0.295 & 0.388 & 0.468 $\pm$ 0.002 & $+0.448\ [0.388,\ 0.508]$ & $\mathbf{1.2\times10^{-10}}$ \\
BinSeg & --- & 0.575 & 0.418 & 0.334 & 0.417 & 0.199 & 0.439 $\pm$ 0.001 & $+0.478\ [0.434,\ 0.523]$ & $\mathbf{1.2\times10^{-10}}$ \\
PELT & --- & 0.497 & 0.345 & 0.222 & 0.149 & 0.181 & 0.330 $\pm$ 0.001 & $+0.586\ [0.537,\ 0.636]$ & $\mathbf{1.2\times10^{-10}}$ \\
ClaSP+KM & --- & 0.333 & 0.171 & 0.101 & 0.068 & 0.038 & 0.187 $\pm$ 0.000 & $+0.730\ [0.692,\ 0.766]$ & $\mathbf{1.2\times10^{-10}}$ \\
\bottomrule
\end{tabular}

}
\end{table*}
\begin{figure}[t]
\centering
\includegraphics[width=0.6\linewidth]{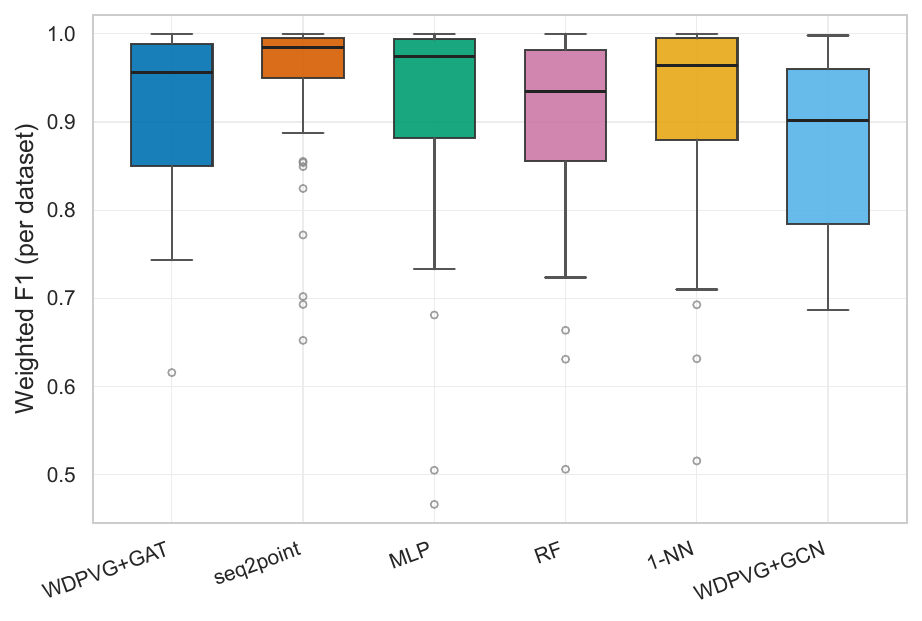}
\caption{ Per-dataset weighted F1 distributions of the supervised methods over the 59 datasets (boxes show quartiles, whiskers $1.5\times$ IQR). WDPVG+GAT is on par with the same-feature non-graph controls and above its attention-free GCN ablation, while the tuned seq2point CNN has the highest median.}
\label{fig:methodbox}
\end{figure}

The ``All'' column of Table~\ref{tab:main_results} reports the weighted F1 of every method, aggregated over the 59 datasets and stratified by segment count. Among the supervised methods the tuned seq2point CNN is the strongest ($0.951$). WDPVG+GAT obtains $0.916$, is statistically indistinguishable from the same-feature MLP, random-forest, and 1-NN controls under Holm correction, and is clearly ahead of its attention-free GCN ablation ($0.877$). The GAT and 1-NN means are identical to three decimals. The unsupervised change-point and state-detection references score lower ($0.19$--$0.68$). Binary Segmentation, Window, and FLUSS receive the ground-truth segment count, while PELT and ClaSP infer change points. The unsupervised methods are deterministic for a fixed series, so their small standard deviations reflect only variation across the five regenerated test sets. These values are contextual because label-free detection and supervised semantic labelling are different tasks. Figure~\ref{fig:methodbox} shows the per-dataset F1 distributions of the supervised methods.

At this benchmark-default resolution, seq2point scores $0.034$ F1 higher than WDPVG+GAT, and the GAT--MLP difference is not significant. WDPVG+GAT uses roughly $70\times$ fewer parameters ($89.6$k vs $6.29$M) and about $46\times$ less peak memory ($73$ vs $3367$\,MB). Section~\ref{sec:resolution} separates two effects. In a common-window native-resolution sweep, the graph is less sensitive to local context width and remains above the same-feature MLP. In a separate resolution sweep that gives seq2point access to windows up to 275 samples, the graph and the window-searched CNN are statistically tied at native resolution rather than the graph holding an accuracy advantage. Seq2point moves ahead after downsampling. This supports an efficiency and robustness interpretation, not a general superiority claim.

\noindent\textbf{Statistical significance.} The planned supervised comparisons in the final column of Table~\ref{tab:main_results} show that WDPVG+GAT is significantly below seq2point (mean gap $-0.034$, CI $[-0.048,-0.020]$, $r=-0.75$, $p_{\mathrm{Holm}}=2.3\times10^{-6}$) and significantly above its matched GCN ablation (mean gap $0.039$, CI $[+0.029,+0.049]$, $r=+0.90$, $p_{\mathrm{Holm}}=9.0\times10^{-9}$). No corrected comparison with a same-feature non-graph control is significant. The MLP gap is $+0.003$ (CI $[-0.014,+0.027]$, $r=-0.11$, $p_{\mathrm{Holm}}=1.00$). The random-forest gap is $+0.012$ (CI $[-0.003,+0.029]$, $r=+0.13$, $p_{\mathrm{Holm}}=1.00$). The 1-NN gap is $+0.0002$ (CI $[-0.014,+0.015]$, median gap $-0.0025$, $r=-0.13$, $p_{\mathrm{Holm}}=1.00$), with GAT higher on 22 and lower on 37 datasets. The GAT architecture improves consistently on the matched GCN. At the downsampled resolution, WDPVG message passing remains statistically indistinguishable from the three same-feature non-graph controls. All five contextual unsupervised references remain below WDPVG+GAT after Holm correction, but these comparisons should not be interpreted as task-equivalent rankings because the methods use no labelled exemplars and require oracle label-permutation alignment for evaluation.

\noindent\textbf{Failure analysis.} The per-dataset performance breakdown is reported in Appendix B. The datasets on which WDPVG+GAT is weakest are \textit{InlineSkate} ($0.62$), \textit{CinCECGTorso} ($0.74$), \textit{ToeSegmentation2} ($0.76$), \textit{FaceFour} ($0.77$) and \textit{MoteStrain} ($0.77$). We hypothesise (consistent with the graph encoding \emph{local waveform shape}) that these are cases in which the segment classes differ little in local shape and are separated instead by longer-range or amplitude-scale properties, or in which long, finely sampled series make a fixed local window span only a small fraction of a segment. A per-class shape/spectral analysis of these datasets is left to future work.

\subsubsection{Performance breakdown across the number of segments} The per-tier columns 3-7 of Table~\ref{tab:main_results} show how each method changes as the number of segments grows from two to six or seven. The ordering is similar across tiers: the tuned seq2point CNN leads at every tier, WDPVG+GAT tracks the same-feature non-graph controls closely, and it stays above its GCN ablation throughout. The tier means are not monotonic, so segment count alone does not explain the variation in supervised performance. Other dataset properties, including waveform separability, may contribute. The failure analysis in Section~\ref{sec:overall} discusses these possibilities cautiously. The unsupervised panel generally weakens as the number of segments increases. ClaSP+KM falls from $0.33$ at two segments to $0.04$ at six or seven, while the change-point baselines vary non-monotonically.

\subsubsection{Attention (GAT vs GCN).} Replacing the GAT layers with matched-capacity GCN layers of the same depth and hidden dimension lowers the mean F1 from $0.916$ to $0.877$ in Table~\ref{tab:main_results}. The paired mean gap is $+0.039$ (95\% CI $[+0.029,+0.049]$), with matched-pairs rank-biserial $r=+0.90$ and Holm-adjusted $p=9.0\times10^{-9}$. The GAT architecture therefore contributes a small but highly consistent gain over the matched GCN. This result supports learned neighbour-specific aggregation, but does not by itself establish which edges receive reduced attention.

\subsubsection{Graph vs same-feature non-graph models.} The gap between the GAT and the MLP, random forest, and 1-NN trained on \emph{identical} windowed features measures the contribution of the graph itself. At the benchmark's downsampled resolution, none of the corrected comparisons with these non-graph controls is significant, as shown in Table~\ref{tab:main_results}. The MLP and random-forest comparisons both give $p_{\mathrm{Holm}}=1.00$, while 1-NN is essentially tied with the GAT (mean gap $+0.0002$, CI $[-0.014,+0.015]$, $r=-0.13$, $p_{\mathrm{Holm}}=1.00$). Against the MLP, which is also evaluated at native resolution, the gap is significant across the tested native-resolution windows, with all Holm-adjusted $p<4\times10^{-6}$, as reported in Section~\ref{sec:resolution}. The graph structure therefore adds resolution-dependent value on top of the features.

\subsubsection{Model robustness to window size, resolution, and efficiency}
\label{sec:resolution}

The aggregate scores of Table~\ref{tab:main_results} are computed at the TSSB benchmark's default temporal resolution, at which the source UCR instances are downsampled (by a median factor of four, and up to $64\times$) before the series are assembled. Downsampling shortens the series and compresses each segment into fewer points. A fixed local window then covers a larger fraction of each segment, which is consistent with the GAT and the windowed non-graph baselines being statistically tied in Table~\ref{tab:main_results}. To probe robustness beyond this single operating point, we vary the input window size at native resolution.

\begin{figure}[t]
\centering
\includegraphics[width=0.5\linewidth]{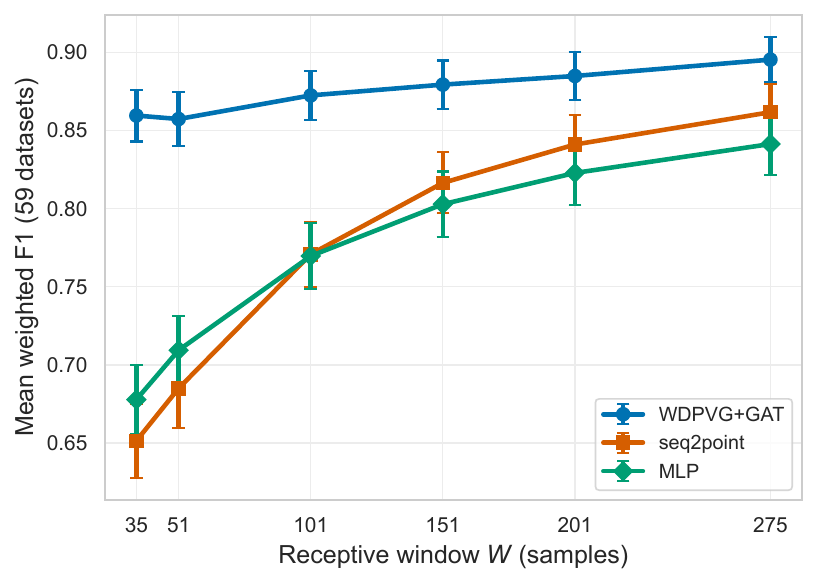}
\caption{\protect Window-size sweep at native resolution (59 datasets). All three methods see a common receptive window $W$, and only the GAT adds the visibility graph. WDPVG+GAT stays near-flat and above both non-graph models at every window, whereas seq2point and the MLP degrade sharply as $W$ shrinks. Error bars show one standard error across datasets.}
\label{fig:window}
\end{figure}

\noindent\textbf{Window-size robustness.} Across all 59 datasets at native (undownsampled) resolution, we sweep a common receptive window $W$ shared by WDPVG+GAT (as node-feature width), seq2point, and the MLP, so that all three receive the same number of raw samples per point and only the GAT additionally uses the visibility graph, as shown in Figure~\ref{fig:window}. The GAT is the least sensitive to the tested local feature-window width: its weighted F1 varies by a within-dataset standard deviation of $0.026$ across the six windows, compared with $0.090$ for seq2point. Its mean F1 exceeds seq2point at every window, with the paired gap increasing from $+0.034$ at $W{=}275$ (GAT higher on $36/59$ datasets, $r=0.39$, $p_{\mathrm{Holm}}=9.0\times10^{-3}$) to $+0.208$ at $W{=}35$ ($57/59$, $r=0.99$, $p_{\mathrm{Holm}}=2.1\times10^{-10}$). The intermediate gaps are $+0.172$, $+0.102$, $+0.063$, and $+0.044$ at $W\in\{51,101,151,201\}$, respectively, and all six comparisons remain significant after Holm correction. The GAT--MLP comparisons are likewise significant at every window. At $W{=}35$, WDPVG+GAT reaches $0.859$ weighted F1, close to seq2point's best global-window result of $0.862$ at $W{=}275$. Performance changes little across the tested context widths. The widening gap at small windows is consistent with the graph providing nonlocal context, but the experiment does not isolate the contribution of long-span edges.

\begin{figure}[t]
\centering
\includegraphics[width=0.6\linewidth]{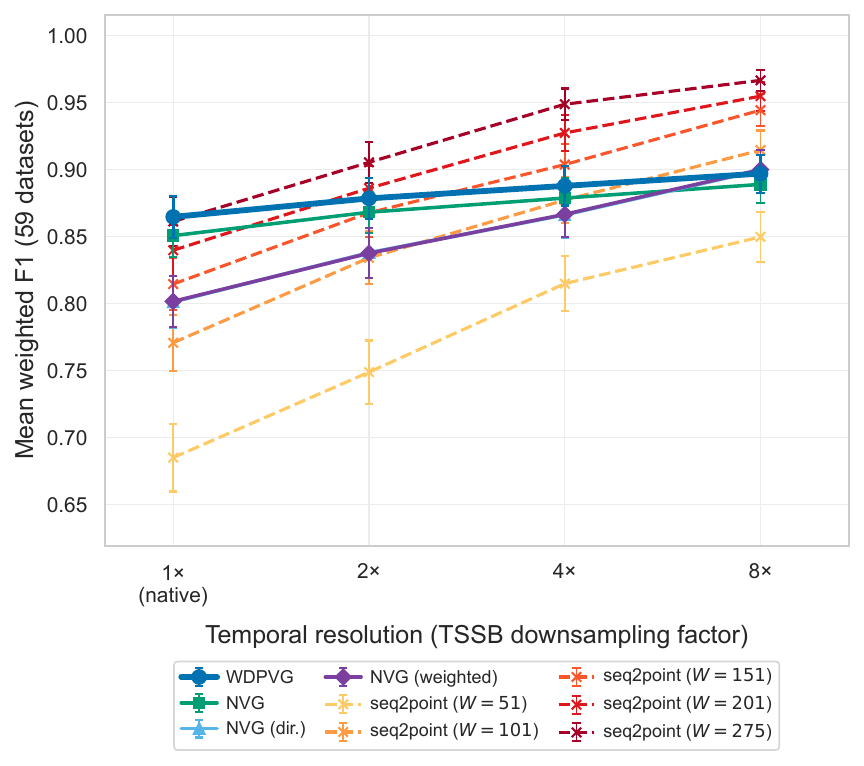}
\caption{Resolution sensitivity of four natural-visibility graph variants and seq2point across its complete tested window grid (59 datasets). WDPVG+GAT and the other graph models keep their fixed node-feature configuration, while the seq2point curves show fixed windows from $W=51$ to $W=275$. At native resolution, WDPVG+GAT and the widest-window seq2point model do not differ significantly ($p_{\mathrm{Holm}}=0.62$), with seq2point higher on $34/59$ datasets. Wider seq2point windows become increasingly advantageous after downsampling. Error bars show one standard error across datasets.}
\label{fig:transform_resolution}
\end{figure}

\noindent\textbf{Resolution and receptive field.} This analysis differs from the common-window sweep above. WDPVG+GAT keeps its fixed validation-selected node-feature window, whereas seq2point is evaluated over $W\in\{51,101,151,201,275\}$ at each temporal resolution. Figure~\ref{fig:transform_resolution} shows the four natural-visibility variants together with the complete seq2point window grid. The grid selects $W=275$, the widest value tested, at every resolution. Across the 59 datasets, WDPVG+GAT and the grid-selected seq2point do not differ significantly at native resolution (mean WDPVG-minus-seq2point gap $+0.003$, median $-0.006$, seq2point higher on $34/59$ datasets, two-sided Wilcoxon $p_{\mathrm{Holm}}=0.62$ with Holm correction across the four resolution contrasts). WDPVG+GAT uses roughly $70\times$ fewer parameters and does not require resolution-specific window selection. After downsampling, seq2point moves ahead: the WDPVG-minus-seq2point gaps are $-0.027$, $-0.061$, and $-0.070$ under $2\times$, $4\times$, and $8\times$ downsampling, with $p_{\mathrm{Holm}}=9.0\times10^{-4}$, $1.4\times10^{-7}$, and $7.4\times10^{-9}$, respectively, so the difference is not significant at native resolution but is significant at every downsampled resolution. These results retract the earlier interpretation that the graph gains an accuracy advantage at native resolution. Once seq2point receives a sufficiently broad receptive-field search, the native comparison shows no significant difference and the main benefit is model size and reduced sensitivity to resolution-specific window selection. Because the largest tested window is selected at every resolution, $W=275$ should be read as the widest value evaluated rather than as a demonstrated optimum.

\begin{table}[t]
\centering
\caption{ Model size and empirical cost (means over the main run). WDPVG+GAT is $\sim\!70\times$ smaller and uses $\sim\!46\times$ less peak memory than seq2point, at the cost of slower full-graph training.}
\label{tab:efficiency}
{
\begin{tabular}{lrrrr}
\toprule
Method & \#Params & Train (s) & Infer (s) & Peak mem (MB) \\
\midrule
WDPVG+GAT & 89.6k & 83.2 & 0.11 & 73 \\
WDPVG+GCN & 87.6k & 40.0 & 0.09 & 35 \\
seq2point & 6.29M & 18.3 & 0.14 & 3367 \\
\bottomrule
\end{tabular}
}
\end{table}

\noindent\textbf{Efficiency.} Table~\ref{tab:efficiency} contrasts model sizes and empirical costs. WDPVG+GAT has $89.6$k parameters (about $70\times$ fewer than the $6.29$M of seq2point) and a $73$\,MB peak-memory footprint roughly $46\times$ smaller in the measured implementation. The graph model does train more slowly per run ($83$\,s vs $18$\,s) in the main benchmark profile, and its attention-free GCN variant roughly halves that cost but at a clear accuracy loss, as shown in Table~\ref{tab:main_results}. Relative to seq2point, the model uses fewer parameters and less measured peak memory, but full-graph training is slower and the default benchmark shows no aggregate accuracy advantage. Its value depends on temporal resolution and available local context.

\subsection{Ablation study results}
\label{sec:results_ablation}

The transform sensitivity was already examined in Section \ref{sec:selecting} while model trade-offs related to attention (GAT vs GCN), graph vs non-graph models, window size, resolution and efficiency were discussed in Section \ref{sec:overall}. In this section we examine the remaining design choices: node features and architecture search. We use a validation-only grid search to examine model capacity and select the final configurations. Every candidate is trained on fitting data and ranked by mean validation weighted F1 across the five stratified 70:30 splits. Parameter count is recorded alongside validation performance, which allows us to identify the non-dominated accuracy--size trade-off.

\begin{figure}[!htbp]
\centering
\includegraphics[width=0.6\linewidth]{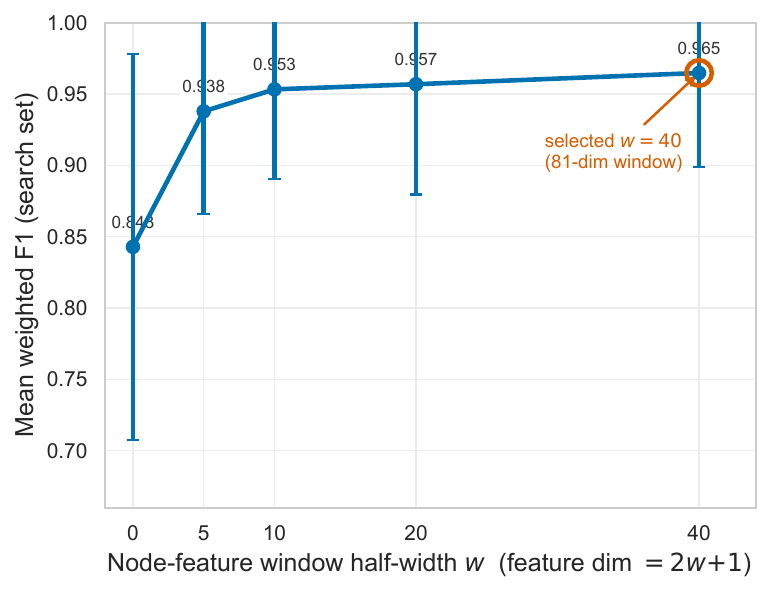}
\caption{Node-feature window sweep during the architecture search
(WDPVG transform). Mean weighted F1 on the search set rises steeply from the scalar node
feature ($w{=}0$) and saturates by $w{=}5$; we select $w{=}40$ (an $81$-dimensional window,
circled) on the plateau. As with the Pareto search (Figure~\ref{fig:pareto}), these are
search-set scores under the search architecture and therefore sit above the $59$-dataset test
scores; the full-benchmark node-feature ablation, where the scalar feature drops to $0.76$, is
reported below. The band is one standard deviation across the
search set.}
\label{fig:featw}
\end{figure}

\noindent\textbf{Node features (windowed vs scalar).} Replacing each node's windowed feature vector by its single scalar value ($w{=}0$) lowers the GAT from $0.916$ to $0.76$ mean weighted F1 in the controlled test ablation over all 59 datasets and five seeds. A scalar node provides little local waveform context, so the model must infer shape only from neighboring node values. The standardised local window supplies the missing local context. The validation-only architecture search shows the same qualitative trend as depicted in Figure \ref{fig:featw}, with mean validation F1 increasing with the window half-width ($0.84$, $0.94$, $0.95$, $0.96$, and $0.965$ for $w\in\{0,5,10,20,40\}$). The architecture-search value of $0.84$ at $w{=}0$ was obtained on the 15-dataset tuning subset with three seeds and the NAS reference architecture. It differs from the controlled 59-dataset ablation in the dataset subset, number of seeds, generated sample counts, and architecture, so the two scalar-feature values are not directly comparable.

\begin{table}[t]
\centering
\caption{ Pareto front of the architecture search (on the WDPVG graph): non-dominated depth/width/heads configurations ordered by parameter count, with mean weighted F1 on the validation search. The selected knee (depth 2, width 32, 8 heads, $w{=}40$, bold) is within $0.0003$ validation F1 of the validation optimum at $7.4\times$ fewer parameters.}
\label{tab:pareto_front_results}
{
\begin{tabular}{ccccrc}
\toprule
Depth & Width & Heads & $w$ & Params & Mean F1 \\
\midrule
2 & 32 & 2 & 40 & 10.1k & 0.953 \\
4 & 32 & 2 & 40 & 18.9k & 0.953 \\
6 & 32 & 2 & 40 & 27.7k & 0.955 \\
2 & 32 & 4 & 40 & 28.3k & 0.961 \\
4 & 32 & 4 & 40 & 62.3k & 0.963 \\
\textbf{2} & \textbf{32} & \textbf{8} & \textbf{40} & \textbf{89.6k} & \textbf{0.967} \\
2 & 96 & 8 & 40 & 661k & 0.967 \\
\bottomrule
\end{tabular}
}
\end{table}
\begin{figure}[t]
\centering
\includegraphics[width=0.6\linewidth]{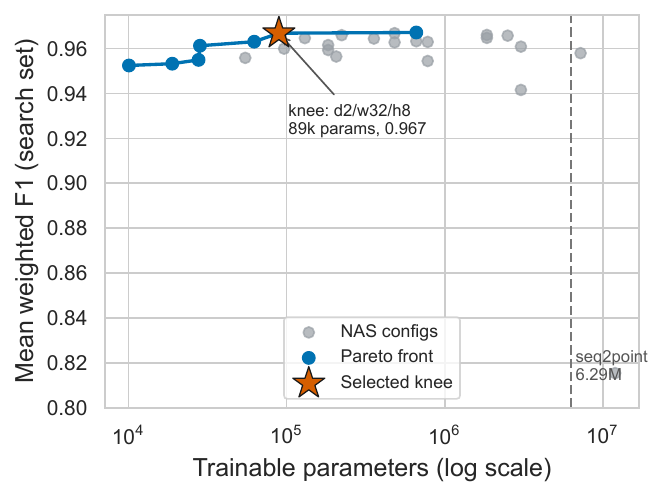}
\caption{Architecture search on fitting/validation data: mean validation weighted F1 against trainable parameter count (log scale) for every configuration in the grid. The Pareto front is highlighted and the selected knee (depth 2, width 32, 8 heads, 89k parameters) lies where validation accuracy saturates. The dashed line marks the seq2point parameter count at 6.29M, about $70\times$ larger.}
\label{fig:pareto}
\end{figure}

\textbf{Architecture search.} The grid covers hidden-layer depth \mbox{$\{2,4,6\}$}, per-head hidden width \mbox{$\{32,96,192\}$}, and attention heads \mbox{$\{2,4,8\}$}. Here, hidden width refers to the output width of each attention head and is distinct from the node-feature half-width $w$. Table~\ref{tab:pareto_front_results} and Figure~\ref{fig:pareto} show the resulting Pareto front. Mean validation F1 is similar at depths \mbox{$2$} and \mbox{$4$} (\mbox{$0.962$} and \mbox{$0.963$}) and falls to \mbox{$0.941$} at depth \mbox{$6$}. The decline is concentrated in deep, wide configurations. At hidden width \mbox{$32$}, the three depths differ by \mbox{$0.002$}. Increasing hidden width from \mbox{$32$} to \mbox{$96$} provides almost no validation gain at the selected depth. We select the Pareto knee with depth \mbox{$2$}, hidden width \mbox{$32$}, eight heads, and feature half-width \mbox{$w{=}40$}. It has \mbox{$89.6$}k parameters and validation F1 \mbox{$0.967$}. The validation-optimal width-96 configuration has \mbox{$661$}k parameters and improves F1 by only \mbox{$0.0003$}. Selection uses mean validation performance over five stratified 70:30 fitting and validation splits generated only from the official UCR \textsc{Train} instances. The UCR \textsc{Test} instances are not used. The selected configuration is then frozen, refitted on series generated from the complete UCR \textsc{Train} pool, and applied unchanged to the official UCR \textsc{Test} instances across all 59 datasets.

\section{Discussion}
\label{sec:discussion}

The graph contribution depends on both temporal resolution and the receptive-field budget of the comparator. At the default downsampled resolution, WDPVG$+$GAT is not significantly different from the same-feature non-graph controls and remains below seq2point. At native resolution, it exceeds the same-feature MLP across the common-window sweep. When seq2point is instead allowed to select a much wider receptive field, the resolution sweep shows a native-resolution tie and a growing seq2point advantage after downsampling, as reported in Section~\ref{sec:resolution}. We next discuss this accuracy--efficiency trade-off, computational cost, failure cases, and limits of the evaluation.

\subsection{When and why the graph representation helps}
\label{subsec:disc_strengths}

The formulation assigns one prediction to each time-point node. The feature window controls the local input context, but it does not control the number or spacing of predictions. The window sweep measures the remaining dependence on that input context.

WDPVG~\cite{zheng2021visibility} connects each point to samples that are visible in the original or reflected series. This produces a polarity-symmetric neighbourhood containing both local and nonlocal edges. In the common-window analysis, this structure lets the compact GAT retain performance as the local feature window narrows and gives it an advantage over the same-feature MLP. The separate resolution analysis shows that a much larger CNN can recover comparable native-resolution accuracy by using a wider window. The experiments therefore support robustness to restricted local context, but they do not isolate individual edge spans or establish an unconditional accuracy advantage over windowed CNNs.

The GAT~\cite{velickovic2018graph} applies neighbour-specific aggregation to the fixed graph. Replacing it with a matched GCN reduces performance consistently, as reported in Section~\ref{sec:results_ablation}. This establishes an architectural difference between GAT and GCN, but does not by itself explain how individual attention coefficients behave. The nonlocal visibility neighbourhood is common to the visibility family. WDPVG additionally represents peaks and troughs through its dual perspective.

The graph model and the MLP use the same windowed node features, so their difference estimates the contribution of graph message passing. That difference is not significant at the downsampled benchmark resolution. At native resolution, the GAT exceeds the same-feature MLP across the tested common windows, as reported in Section~\ref{sec:resolution}. The gap is largest when the local feature window covers only a small part of a segment. This result is consistent with useful nonlocal graph context. It remains distinct from the seq2point comparison, in which the CNN is allowed a broader, resolution-specific receptive-field search. A dedicated edge-span ablation would still be needed to identify the responsible connections.

\subsection{Shape generalisation across unseen instances}
\label{subsec:disc_cnn}

Each final test series is assembled from UCR instances that are not used for model development. This prevents direct reuse of the same raw instances across development and test data. Models must transfer class information across instance-level variation in amplitude, offset, phase, and noise. The natural-visibility edge set is invariant to affine transformations of the plotted series~\cite{lacasa2008time}, which may reduce sensitivity to some amplitude and offset changes. Visibility-graph GNNs have also performed well in whole-series UCR classification~\cite{xuan2022clpvg,coelho2024alchemist,cheng2023time2graph}. The present results extend this line of work to dense per-point labelling. They show that graph message passing is most useful when local context is restricted, while a sufficiently wide CNN can match its native-resolution accuracy.

\subsection{Scope: relationship to change-point and state-detection methods}
\label{subsec:disc_scope}

The present task uses labelled training instances and assigns a known semantic class to every test time point. ClaSP and CLaP instead detect changes without labelled exemplars and derive states from the resulting segments~\cite{ermshaus2023clasp,ermshaus2025clap}. ClaSP+KM is included only as an unsupervised reference and is scored with the same per-point metric after label alignment. Its lower score should not be read as a direct ranking against supervised semantic labelling. The two settings answer different questions: one identifies distributional changes without labels, and the other recognises known segment classes from labelled examples.

\subsection{Computational cost and scalability}
\label{subsec:disc_cost}

Table~\ref{tab:transform_efficiency} shows that graph size is the main computational constraint. Naive visibility-graph construction costs $\mathcal{O}(N^2)$, while divide-and-conquer algorithms can reduce this to $\mathcal{O}(N\log N)$. WDPVG constructs the graph for both the original and reflected series, which increases the constant cost without changing the asymptotic order~\cite{silva2021time}. Window-feature extraction is linear. A GAT epoch costs $\mathcal{O}\!\big(L(Ed+Nd^2)\big)$, where $E$ is the number of stored edges and $d$ is the hidden dimension~\cite{velickovic2018graph}.

The edge count depends on the signal. Smooth or monotone regions can produce dense natural-visibility graphs, making memory and message passing expensive for long series. Horizontal or limited-penetrable visibility graphs, neighbourhood sampling, mini-batched message passing, and sparse-attention methods offer possible reductions in cost~\cite{luque2009horizontal,zhou2012limited}. These alternatives trade graph detail for scalability.

The transform measurements show the same cost pattern. WDPVG uses $73.1$~MB and $6{,}348$ stored message-passing edges per test graph on average, approximately twice the peak memory of directed NVG ($31.7$~MB) without an F1 gain. Conversely, horizontal-visibility graphs use only $22$--$25$~MB. The GAT is parameter-efficient relative to seq2point. WDPVG itself is not the most memory-efficient visibility representation.

\subsection{Hypothesised failure mechanisms}
\label{subsec:disc_failure}

The lower-scoring datasets motivate several hypotheses for future analysis. Classes may differ mainly in spectral content or amplitude rather than in visible waveform shape. Some datasets also contain few training instances per class, which limits cross-instance learning. Smooth or monotone signals can create dense graphs that increase cost and spread aggregation over large neighbourhoods. Finally, a visibility graph may be poorly matched to boundary changes that do not alter the visible shape of the signal. These explanations have not been isolated experimentally and should be treated as hypotheses rather than established failure modes.

\subsection{Generalisation to unseen instances and domains}
\label{subsec:disc_generalization}

Training and test series are built from disjoint UCR instances, so no raw instances, graph nodes, or graph edges are shared between them. The 59 TSSB-derived datasets span electrocardiography and other physiological signals, gesture and motion capture, appliance and robot sensors, image-outline profiles, and food or material spectra. The study nevertheless does not test transfer between domains. A separate model is trained for each dataset. The protocol also preserves the fixed TSSB segment order and does not test unseen permutations. Finally, it evaluates unseen instances of known classes rather than entirely new classes. Pretraining a graph encoder across datasets is one possible extension towards transferable representations~\cite{ermshaus2024class}.

\subsection{Towards multivariate time series}
\label{subsec:disc_multivariate}

A multivariate extension could follow either of two designs. A multilayer visibility graph would build one graph per channel and connect contemporaneous nodes across channels~\cite{silva2025mhvg,silva2025mqg}. This preserves the time-point node representation but increases the edge count. Another option is to represent variables as nodes and learn their adjacency, as in MTGNN, TodyNet, and MTAD-GAT~\cite{wu2020mtgnn,liu2024todynet,zhao2020mtadgat}. Both options require additional work on sparsification and scalability. Whole-series classification and regime characterisation are also natural applications of the same graph representation~\cite{he2024adaptive,peng2025vector}.

\section{Conclusions}
\label{sec:conclusions}

We formulated dense univariate time-series segmentation as supervised node classification. Each time point is represented by one graph node and receives one output label. A local window provides input features but does not set the output spacing. The TSSB-derived inductive benchmark separates development instances from the UCR test instances and therefore evaluates generalisation to unseen instances.

Under one fixed GAT architecture, the visibility-based transformations achieve the highest mean ranks among the eleven graph constructions tested. The visibility variants are not all equivalent. WDPVG, directed NVG, and weighted NVG form a statistically indistinguishable top group after Holm correction. WDPVG is used for its polarity-symmetric construction, not because it is uniquely more accurate or more efficient. The matched GAT performs consistently better than the matched GCN when both receive the same graph topology and edge weights.

The results define a clear operating regime for the graph approach. At the default downsampled resolution, WDPVG+GAT reaches weighted F1 $0.916$, below seq2point at $0.951$ and statistically indistinguishable from the same-feature MLP, random forest, and 1-NN controls. For finely sampled native-resolution series, however, it is less sensitive to the feature-window width and significantly outperforms the same-feature MLP at every tested window. In the resolution sweep, it also matches a window-searched seq2point at native resolution while using about $70\times$ fewer parameters and $46\times$ less measured peak memory. Seq2point moves ahead after downsampling, and graph construction and full-graph training add cost. WDPVG+GAT is therefore best viewed as a compact segmenter for settings where native temporal detail, model size, and the cost of repeated window tuning matter more than maximising accuracy after aggressive downsampling.

\section*{Funding}

This work was supported by the Slovenian Research Agency (P2-0016) and the European Commission NANCY (No. 101096456), AcceptRemote (No. 101269278) and 6G-OPTICON (No. 101292327) projects.

\section*{Author contributions statement}

I.K. and B.B. conceived the experiments. I.K. conducted the experiments. B.B. and I.K. analysed the results. M.M. and C.F. acquired funding. I.K. and B.B. wrote the original draft. All authors reviewed the manuscript. 

\bibliography{sample}

\section*{Appendix A: Time series to graph transformations}
\label{app:transformations}
This appendix gives the formal definitions of the transformation families introduced in Section~\ref{sec:transf}. Table~\ref{tab:network_comparison} lists all variants evaluated in this work, together with their node representation, edge type, and whether the resulting network is directed and weighted.

\begin{table}[h!]
\centering
\renewcommand{\arraystretch}{1.8}
\setlength{\tabcolsep}{8pt}
\begin{tabular}{|c|l|c|c|c|c|}
\hline
\multirow{8}{*}{{\textbf{\ \ \ \  VISIBILITY}}} & \textbf{Network type}            & \textbf{node} & \textbf{edge} & \textbf{directed} & \textbf{weighted} \\ \cline{2-6}
& Natural Visibility Graph (NVG)                            & t             & NV            & $\times$          & $\times$          \\ \cline{2-6}
& Horizontal Visibility Graph (HVG)                           & t             & HV            & $\times$          & $\times$          \\ \cline{2-6}
& Directed NVG                    & t             & NV            & \checkmark        & $\times$          \\ \cline{2-6}
& Directed HVG                    & t             & HV            & \checkmark        & $\times$          \\ \cline{2-6}
& Weighted NVG                    & t             & NV            & \checkmark        & \checkmark        \\ \cline{2-6}
& Weighted HVG                    & t             & HV            & \checkmark        & \checkmark        \\ \cline{2-6}
& Weighted Dual Perspective Visibility Graph            & t             & NV            & $\times$          & \checkmark        \\ \cline{2-6}
\hline
\multirow{3}{*}{{\textbf{\ \ \ \  TRANSITION}}} & Quantile Network                & $t_i\,(q_i)$  & TP            & \checkmark        & \checkmark        \\ \cline{2-6}
& Ordinal Partition Transition Network                            & $t_i\,(\pi_i)$ & TP            & \checkmark        & \checkmark        \\ \cline{2-6}
& Coarse-Grained Phase Space Graph                           & $t_i\,(\vec{z}_i)$ & TP            & \checkmark        & \checkmark        \\ \hline
\textbf{PROXIMITY}                & k-NN Network                    & $t_i\,(\vec{z}_i)$ & DM            & \checkmark        & \checkmark        \\ \hline
\end{tabular}
\caption{Time-series-to-graph transformations evaluated in this work.}
\label{tab:network_comparison}
\end{table}

\subsection*{Visibility networks}

Visibility transformations follow the classical visibility algorithms of computational geometry~\cite{ghosh2007visibility} and respect the temporal order of the series. The Natural Visibility Graph (NVG) of Lacasa et al.~\cite{lacasa2008time} treats each data point as a vertical bar whose height equals its value. Two points $(t_i,s_i)$ and $(t_j,s_j)$ are connected if every intermediate point $(t_k,s_k)$ satisfies

\begin{equation*}
s_k < s_i + \frac{(s_j - s_i)(t_k - t_i)}{(t_j - t_i)}, \quad i < k < j.
\end{equation*}

The edges form an adjacency matrix $A^{NVG}$. The NVG is simple to define and has quadratic complexity $O(N^2)$ under the direct construction~\cite{silva2021time}. A single upper-perspective NVG does not represent peaks and troughs symmetrically, and a formulation based only on sample indices does not encode irregular time intervals explicitly.

The Weighted Dual-Perspective Visibility Graph (WDPVG)~\cite{zheng2021visibility} addresses both issues. In this study, an edge between $(t_i,s_i)$ and $(t_j,s_j)$ is weighted by their Euclidean distance in the time--value plane,
\begin{equation*}
w_{ij}=\sqrt{(t_j-t_i)^2+(s_j-s_i)^2}.
\end{equation*}
A second weighted NVG is built on the reflected series $S'=-S$. Because Euclidean distance is invariant to this sign reflection, the dual perspective changes the set of visible edges, while shared edges retain the same weight. The two weighted adjacency matrices are combined edge by edge,

\begin{equation*}
A^{WDPVG}_{i,j} = \max\!\left(A^{NVG}_{i,j},\, A^{RPNVG}_{i,j}\right),
\end{equation*}

so that connections visible from either perspective are kept. The nonzero weights are then divided by the maximum weight within that graph, giving retained edge values satisfying $0<w_{ij}\leq 1$. The implementation stores the resulting graph as an undirected message-passing graph by including both edge orientations. The nodes $v_n$ are the measurements $s_n$, and the edges are stored in $A^{WDPVG}$.

The Horizontal Visibility Graph (HVG)~\cite{luque2009horizontal} uses a stricter, horizontal criterion: two nodes $v_i$ and $v_j$ are connected only if every intermediate node $v_k$ satisfies

\begin{equation*}
s_i, s_j > s_k, \quad \forall k \in (i, j).
\end{equation*}

It is simpler and faster than the NVG but captures less structure.

Weighted~\cite{supriya2016weighted} and directed~\cite{lacasa2012time} visibility graphs add further structure. A direction can be assigned to each edge, either left-to-right along the temporal axis or top-to-bottom along the value axis, and each edge can carry a weight such as the Euclidean or squared Euclidean distance, the horizontal or vertical distance, or the slope between the connected points.

\subsection*{Transition networks}

Transition-based mappings first encode each time point by a discrete or reconstructed state while preserving the time-indexed node set $V=\{v_1,\ldots,v_N\}$. Let $q_i=f(s_i)\in Q$ denote the state assigned to time point $i$. A first-order state-transition matrix $P$ is estimated from consecutive encoded points,

\begin{equation*}
P_{ab}=\Pr(q_{i+1}=b\mid q_i=a), \qquad a,b\in Q.
\end{equation*}

The state-level probabilities are then lifted back to the time axis as a Markov transition field,

\begin{equation*}
A^{\mathrm{MTF}}_{ij}=P_{q_i q_j}, \qquad i,j\in\{1,\ldots,N\}.
\end{equation*}

Node $v_i$ still represents the original time point $s_i$. Its state $q_i$ determines the transition-valued relationships to other time-point nodes. The construction supports one node label per sample, as required by the dense-segmentation formulation. Quantile, ordinal-pattern, and coarse-grained phase-space variants differ in the encoding $f$ and the corresponding state sequence. Because the lifted field can contain $O(N^2)$ relations, it is costly for long series unless sparsified.

\subsection*{Proximity networks}

Proximity networks embed the series in a multidimensional phase space and connect points according to a distance $d(s_i, s_j)$ or similarity $r(s_i, s_j)$ measure, as in the k-nearest-neighbour network. The adjacency matrix can be written as:

\begin{equation*}
A_{i,j} =
\begin{cases}
1 & \text{if } d(s_i, s_j) \leq \epsilon \text{ or } r(s_i, s_j) \geq \alpha, \\
w_{i,j} & \text{if a weighting is applied}, \\
0 & \text{otherwise}.
\end{cases}
\end{equation*}

This family includes cycle~\cite{zhang2006complex, zhang2006detecting}, correlation~\cite{yang2008complex}, and recurrence~\cite{marwan2008historical} networks.

\section*{Appendix B: Detailed segmentation results}

Table~\ref{tab:results-classification} reports the per-dataset weighted F1 of the proposed WDPVG+GAT and the four supervised baselines (seq2point, MLP, random forest, and 1-NN) using the setup of Section~\ref{sec:expsetup}, averaged over the five seeds and ordered by the number of segments. The final two rows give the mean and standard deviation across all 59 datasets. Consistent with Table~\ref{tab:main_results}, seq2point leads on average while WDPVG+GAT tracks the non-graph controls.

{
\footnotesize
\begin{longtable}{lcccccc}
\caption{Per-dataset weighted F1 (mean over five seeds), ordered by segment count.}\label{tab:results-classification}\\
\toprule
Dataset & Seg. & WDPVG+GAT & seq2point & MLP & RF & 1-NN \\
\midrule
\endfirsthead
\toprule
Dataset & Seg. & WDPVG+GAT & seq2point & MLP & RF & 1-NN \\
\midrule
\endhead
ArrowHead & 2 & 0.987 & 0.998 & 0.994 & 0.961 & 0.998 \\
Beef & 2 & 0.958 & 0.985 & 0.927 & 0.920 & 0.922 \\
BeetleFly & 2 & 0.779 & 0.702 & 0.733 & 0.738 & 0.710 \\
BirdChicken & 2 & 0.812 & 0.887 & 0.762 & 0.740 & 0.780 \\
ChlorineConcentration & 2 & 0.975 & 0.970 & 0.977 & 0.993 & 0.993 \\
Coffee & 2 & 0.998 & 1.000 & 1.000 & 1.000 & 1.000 \\
Computers & 2 & 0.823 & 0.946 & 0.815 & 0.856 & 0.711 \\
DistalPhalanxOutlineAgeGroup & 2 & 0.987 & 0.999 & 1.000 & 1.000 & 1.000 \\
ECG200 & 2 & 0.977 & 0.984 & 0.993 & 0.991 & 0.993 \\
ECGFiveDays & 2 & 0.971 & 0.991 & 0.996 & 0.835 & 0.996 \\
GunPoint & 2 & 0.990 & 0.994 & 0.994 & 0.965 & 0.993 \\
ItalyPowerDemand & 2 & 0.994 & 0.995 & 0.999 & 0.997 & 1.000 \\
Lightning2 & 2 & 0.956 & 0.987 & 0.993 & 0.948 & 0.980 \\
MedicalImages & 2 & 0.985 & 0.993 & 0.987 & 0.970 & 0.986 \\
MoteStrain & 2 & 0.771 & 0.899 & 0.813 & 0.832 & 0.849 \\
SonyAIBORobotSurface1 & 2 & 0.949 & 0.971 & 0.914 & 0.912 & 0.950 \\
SonyAIBORobotSurface2 & 2 & 0.882 & 0.974 & 0.873 & 0.885 & 0.900 \\
ToeSegmentation1 & 2 & 0.803 & 0.939 & 0.767 & 0.823 & 0.777 \\
ToeSegmentation2 & 2 & 0.760 & 0.849 & 0.770 & 0.789 & 0.822 \\
TwoLeadECG & 2 & 0.990 & 0.991 & 0.996 & 0.931 & 0.996 \\
Yoga & 2 & 0.975 & 0.996 & 0.980 & 0.979 & 0.974 \\
CBF & 3 & 0.951 & 0.985 & 0.976 & 0.905 & 0.975 \\
CinCECGTorso & 3 & 0.743 & 0.693 & 0.681 & 0.631 & 0.693 \\
DiatomSizeReduction & 3 & 1.000 & 0.986 & 0.976 & 0.979 & 0.997 \\
DistalPhalanxTW & 3 & 0.988 & 1.000 & 1.000 & 1.000 & 1.000 \\
Haptics & 3 & 0.963 & 0.980 & 0.837 & 0.862 & 0.800 \\
InlineSkate & 3 & 0.616 & 0.652 & 0.505 & 0.506 & 0.516 \\
LargeKitchenAppliances & 3 & 0.792 & 0.916 & 0.890 & 0.820 & 0.632 \\
Meat & 3 & 0.999 & 1.000 & 0.992 & 0.993 & 1.000 \\
OSULeaf & 3 & 0.920 & 0.990 & 0.949 & 0.955 & 0.902 \\
OliveOil & 3 & 0.995 & 0.855 & 0.466 & 0.864 & 0.928 \\
ProximalPhalanxOutlineAgeGroup & 3 & 1.000 & 0.999 & 1.000 & 1.000 & 1.000 \\
Trace & 3 & 0.996 & 0.999 & 0.991 & 0.935 & 0.964 \\
WordSynonyms & 3 & 0.914 & 0.946 & 0.938 & 0.724 & 0.908 \\
Adiac & 4 & 0.983 & 0.986 & 0.974 & 0.958 & 0.947 \\
Car & 4 & 0.974 & 0.989 & 0.961 & 0.842 & 0.941 \\
FaceFour & 4 & 0.771 & 0.854 & 0.767 & 0.795 & 0.809 \\
FiftyWords & 4 & 0.886 & 0.982 & 0.932 & 0.958 & 0.898 \\
InsectWingbeatSound & 4 & 0.910 & 0.987 & 0.987 & 0.922 & 0.995 \\
Mallat & 4 & 0.991 & 0.997 & 0.957 & 0.961 & 0.933 \\
ProximalPhalanxTW & 4 & 0.997 & 0.997 & 1.000 & 0.983 & 1.000 \\
SwedishLeaf & 4 & 0.999 & 0.999 & 0.998 & 1.000 & 0.997 \\
Symbols & 4 & 0.949 & 0.956 & 0.939 & 0.903 & 0.978 \\
CricketX & 5 & 0.827 & 0.976 & 0.900 & 0.855 & 0.882 \\
CricketY & 5 & 0.869 & 0.984 & 0.943 & 0.935 & 0.949 \\
CricketZ & 5 & 0.832 & 0.972 & 0.916 & 0.888 & 0.874 \\
Lightning7 & 5 & 0.816 & 0.772 & 0.765 & 0.664 & 0.744 \\
ShapesAll & 5 & 0.918 & 0.927 & 0.926 & 0.922 & 0.929 \\
SyntheticControl & 5 & 0.959 & 0.999 & 1.000 & 0.999 & 0.999 \\
UWaveGestureLibraryAll & 5 & 0.952 & 0.993 & 0.982 & 0.966 & 0.993 \\
UWaveGestureLibraryX & 5 & 0.896 & 0.974 & 0.990 & 0.998 & 0.989 \\
UWaveGestureLibraryY & 5 & 0.911 & 0.979 & 0.989 & 0.990 & 0.994 \\
UWaveGestureLibraryZ & 5 & 0.874 & 0.979 & 0.978 & 0.960 & 0.987 \\
NonInvasiveFetalECGThorax1 & 6 & 0.993 & 0.999 & 0.997 & 0.989 & 0.995 \\
FaceAll & 7 & 0.826 & 0.954 & 0.812 & 0.886 & 0.878 \\
FacesUCR & 7 & 0.780 & 0.825 & 0.772 & 0.826 & 0.814 \\
Fish & 7 & 0.964 & 0.964 & 0.916 & 0.926 & 0.902 \\
NonInvasiveFetalECGThorax2 & 7 & 0.997 & 0.999 & 0.996 & 0.980 & 0.992 \\
Plane & 7 & 0.999 & 0.999 & 1.000 & 1.000 & 1.000 \\
\midrule
\textbf{Mean} & & 0.916 & 0.951 & 0.913 & 0.904 & 0.916 \\
\textbf{STD} & & 0.089 & 0.080 & 0.118 & 0.104 & 0.108 \\
\bottomrule
\end{longtable}
}


\end{document}